\documentclass{article}

\usepackage[nonatbib, final]{neurips_2022}

\usepackage[square,numbers, sort&compress]{natbib}
\usepackage[utf8]{inputenc} %
\usepackage[T1]{fontenc}    %
\usepackage{hyperref}       %
\usepackage{url}            %
\usepackage{booktabs}       %
\usepackage{amsfonts}       %
\usepackage{nicefrac}       %
\usepackage{microtype}      %
\usepackage{xcolor}         %

\usepackage{amssymb, amsmath, mathtools}
\usepackage{amsthm,amssymb}
\usepackage{adjustbox} 
\usepackage{comment}
\usepackage{enumitem} 

\usepackage{algorithm}
\usepackage[noend]{algpseudocode}
\usepackage{booktabs}
\usepackage{wrapfig}

\DeclareMathOperator*{\argmax}{arg\,max}
\DeclareMathOperator*{\argmin}{arg\,min}
\DeclareMathOperator{\E}{\mathop{\mathbb{E}}}
\DeclareMathOperator{\R}{\mathop{\mathbb{R}}}
\newcommand{\g}{\nabla_\theta}

\newcommand*{\annot}[1]{\tag*{\footnotesize{\textcolor{black!50}{#1}}}}

\definecolor{bostonuniversityred}{rgb}{0.8, 0.0, 0.0}
\newcommand{\red}[1]{{\color{bostonuniversityred}#1}}
\definecolor{ao}{rgb}{0.0, 0.5, 0.0}
\newcommand{\green}[1]{{\color{ao}#1}}

\DeclarePairedDelimiter\ceil{\lceil}{\rceil}
\DeclarePairedDelimiter\floor{\lfloor}{\rfloor}

\newtheorem{theorem}{Theorem}[section]

\newtheorem{lemma}[theorem]{Lemma}

\title{Policy Gradient With Serial Markov Chain Reasoning}

\author{%
  Edoardo Cetin \\
  Department of Engineering\\
  King's College London \\
  \texttt{edoardo.cetin@kcl.ac.uk} \\
   \And
   Oya Celiktutan \\
   Department of Engineering\\
  King's College London \\
  \texttt{oya.celiktutan@kcl.ac.uk} \\
}

\begin{document}

\maketitle

\begin{abstract}

We introduce a new framework that performs decision-making in reinforcement learning (RL) as an iterative \textit{reasoning} process. We model agent behavior as the \textit{steady-state distribution} of a parameterized \textit{reasoning} Markov chain (RMC), optimized with a new tractable estimate of the policy gradient. We perform action selection by simulating the RMC for enough \textit{reasoning steps} to approach its steady-state distribution. We show our framework has several useful properties that are inherently missing from traditional RL. For instance, it allows agent behavior to approximate any continuous distribution over actions by parameterizing the RMC with a simple Gaussian transition function. Moreover, the number of reasoning steps to reach convergence can scale adaptively with the difficulty of each action selection decision and can be accelerated by re-using past solutions. Our resulting algorithm achieves state-of-the-art performance in popular Mujoco and DeepMind Control benchmarks, both for proprioceptive and pixel-based tasks.

\end{abstract}

\section{Introduction}

\label{sec:intro}

Reinforcement learning (RL) has the potential to provide a general and effective solution to many modern challenges. Recently, this class of methods achieved numerous impressive milestones in different problem domains, such as games \citep{dqn, alphazero, alphastar}, robotics \citep{qt-opt, rubics_cube, robotic-stacking-shape}, and other meaningful real-world applications \citep{success-balloons, success-tokamak, success-youtube-muzero}. However, all these achievements relied on massive amounts of data, controlled environments, and domain-specific tuning. These commonalities highlight some of the current practical limitations that prevent RL to be widely applicable \citep{challenges-rl}.

\begin{wrapfigure}{r}{0.4\textwidth}
\vspace{-9mm}
\begin{center}
    \includegraphics[width=0.4\textwidth]{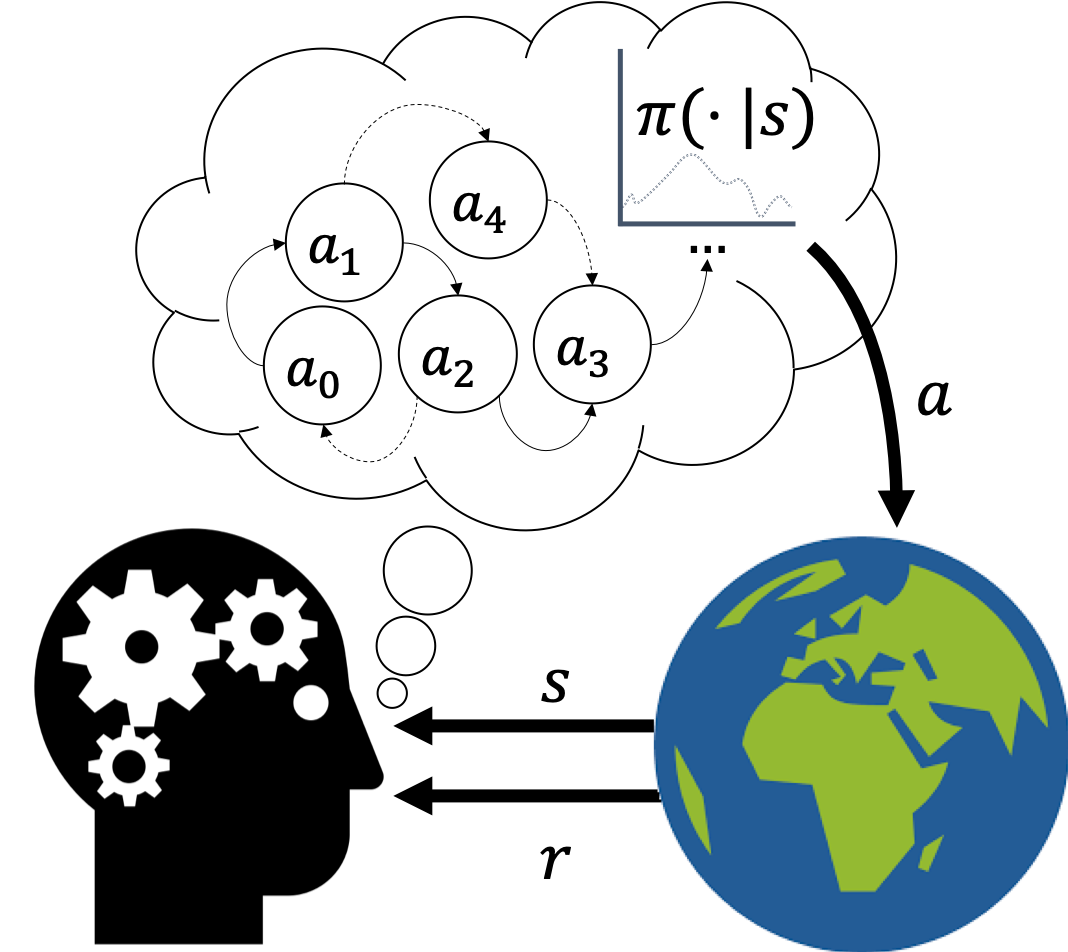}
  \end{center}
  \vspace{-4mm}
  \caption{\small Depiction of agent decision-making with
  serial Markov chain reasoning. }
  \label{fig:res_vis}
  \vspace{-4mm}
\end{wrapfigure}

In the deep RL framework, practitioners train agents with the end goal of obtaining optimal \textit{behavior}. Traditionally, agent behavior is modeled with \textit{feed-forward policies} regressing from any state to a corresponding distribution over actions. Such formulation yields practical training objectives in both off-policy \citep{dpg,ddpg,sac} and on-policy settings \citep{a3c,ppo,ppg}. However, we identify three inherent properties of this rigid representation of behavior that could considerably impact expressivity and efficiency in continuous control tasks. First, agent behavior is restricted to a class of tractable distributions, which might fail to capture the necessary complexity and multi-modality of a task. Second, the policy performs a fixed \textit{reasoning} process with a feed-forward computation, which potency \textit{cannot adapt} to the varying complexity of individual action selection problems. Third, decision-making is performed every time \textit{from scratch}, without re-using any past information that might still inform and facilitate the 
current action selection problem.

Unlike RL policies, human reasoning does not appear to follow a rigid feed-forward structure. In fact, a range of popular psychological models characterize human decision-making as a \textit{sequential} process with adaptive temporal dynamics \citep{cogn-poissSem01966, cogn-diffSEM01978, cogn-accSEM02001, cogn-linbaccSEM2008}. Many of these models have found empirical groundings in neuroscience \citep{cogn-diffNeuro02001inprimates, cogn-diffNeuro12011, cogn-diffNeuro22012, cogn-diffNeuro32016} and have shown to effectively complement RL for capturing human behavior in experimental settings \citep{cogn-diffRL2017, cogn-diffRL2019}. Partly inspired by these works, we attempt to \textit{reframe} the deep RL framework by making use of a similar flexible model of agent behavior, in order to counteract its aforementioned limitations.

We introduce \textit{serial Markov chain reasoning} \-- a new powerful framework for representing agent behavior. Our framework treats decision-making as an \textit{adaptive reasoning process}, where the agent sequentially updates its beliefs regarding which action to execute in a series of \textit{reasoning steps}. We model this process by replacing the traditional policy with a parameterized \textit{transition function}, which defines a \textit{reasoning Markov chain} (RMC). The steady-state distribution of the RMC represents the distribution of agent behavior after performing enough \textit{reasoning} for decision-making. Our framework naturally overcomes the aforementioned limitations of traditional RL. In particular, we show that our agent's behavior can approximate any arbitrary distribution even with simple parameterized transition functions. 
Moreover, the required number of \textit{reasoning steps} adaptively scales with the difficulty of individual action selection problems and can be accelerated by re-using samples from similar RMCs.

To optimize behavior modeled by the steady-state distribution of the RMC, we derive a new tractable method to estimate the \textit{policy gradient}. Hence, we implement a new effective off-policy algorithm for maximum entropy reinforcement learning (MaxEnt RL) \citep{maxentobj, rlasinf6-levineTut}, named \textit{Steady-State Policy Gradient} (SSPG). Using SSPG, we empirically validate the conceptual properties of our framework over traditional MaxEnt RL. Moreover, we obtain state-of-the-art results for popular benchmarks from the OpenAI Gym Mujoco suite \citep{gym} and the DeepMind Control suite from pixels \citep{dmc}.

In summary, this work makes the following key contributions:
\begin{enumerate}[leftmargin=*] %
    \item We propose \textit{serial Markov Chain reasoning}  a framework to represent agent behavior that can overcome expressivity and efficiency limitations inherent to traditional reinforcement learning.
    \item Based on our framework, we derive \textit{SSPG}, a new tractable off-policy algorithm for MaxEnt RL.
    \item We provide experimental results validating theorized properties of \textit{serial Markov Chain reasoning} and displaying state-of-the-art performance on the Mujoco and DeepMind Control suites.
\end{enumerate}

\section{Background}

\label{sec:background}

\subsection{Reinforcement learning problem}

We consider the classical formulation of the reinforcement learning (RL) problem setting as a Markov Decision Process (MDP) \cite{mdp}, defined by the tuple $(S, A, P, p_0, r, \gamma)$. In particular, at each discrete time step $t$ the agent experiences a state %
from the environment's state-space, $s_t\in S$, based on which it selects an action from its own action space, $a_t\in A$. In continuous control problems (considered in this work), the action space is typically a compact subset of an Euclidean space $\R^{dim(A)}$. The evolution of the environment's state through time is determined by the transition dynamics and initial state distribution, $P$ and $p_0$. Lastly, the reward function $r$ represents the immediate level of progress for any state-action tuple towards solving a target task. %
The agent's behavior is represented by a state-conditioned parameterized policy  distribution $\pi_\theta$. Hence, its interaction with the environment produces trajectories, $\tau = (s_0, a_0, s_1, ..., s_T, a_T)$, according to a factored joint distribution
$p_{\pi_{\theta}}(\tau)=p_0(s_0)\prod^{T}_{t=0}\pi_\theta(a_t|s_t)P(s_{t+1}|s_t, a_t)$. The RL objective is to optimize agent behavior as to maximize the discounted sum of expected future rewards: $\argmax_\theta \E_{p_{\pi_{\theta}}(\tau)}\left[ \sum^{T}_{t = 0}\gamma^t r (s_t, a_t) \right]$.

\subsection{Maximum entropy reinforcement learning and inference}

Maximum entropy reinforcement learning (MaxEnt RL) \citep{maxentirl} considers optimizing agent behavior for a different objective that naturally arises when formulating action selection as an inference problem \citep{rlasinf2-planning, rlasinf3-todorovduality, rlasinf4-mdp, rlasinf5-maxentziebart}. Following \citet{rlasinf6-levineTut}, we consider modeling a set of binary optimality random variables with realization probability proportional to the exponentiated rewards scaled by the temperature $\alpha$, $p(O_t|s_t, a_t) \propto \exp(\frac{1}{\alpha}r(s_t, a_t))$. The goal of MaxEnt RL is to minimize the KL-divergence between trajectories stemming from agent behavior, $p_{\pi_{\theta}}(\tau)$, and the inferred optimal behavior, $p(\tau|O_{0:T})$:
\begin{equation} \label{DKLOBJ}
\begin{split}
    D_{KL}\left(p_{\pi_{\theta}}(\tau)||p(\tau|O_{0:T})\right) &= \E_{p_{\pi_{\theta}(\tau)}}\left[\log \frac{p_0(s_0)\prod^{T}_{t=0}\pi_\theta(a_t|s_t)P(s_{t+1}|s_t, a_t)}{p_0(s_0)\prod^{T}_{t=0} \exp(\frac{1}{\alpha}r(s_t, a_t))P(s_{t+1}|s_t, a_t)}\right] \\
     &=-\E_{p_{\pi_{\theta}}(\tau)}\left[ \sum^{T}_{t = 0} r (s_t, a_t) + \alpha H(\pi(\cdot|s_t))\right].
\end{split}
\end{equation}
The resulting entropy-regularized objective introduces an explicit trade-off between exploitation and exploration, regulated by the temperature parameter $\alpha$ scaling the policy's entropy. An effective choice to optimize this objective %
is to learn an auxiliary parameterized \textit{soft Q-function} \citep{softqfunction}:
\begin{equation} \label{eqn:soft-Q}
    Q_\phi^\pi(s_t, a_t) = \E_{p_{\pi_{\theta}}(\tau|s_t, a_t)}\left[ r(s_t, a_t) + \sum^{T}_{t' = {t+1}} r (s_{t'}, a_{t'}) + \alpha H(\pi(a_{t'}|s_{t'})\right].
\end{equation}
Given some state, $Q_\phi^\pi(s, \cdot)$ represents an energy-function based on the expected immediate reward and the agent's future likelihood of optimality from performing any action. Thus, we can locally optimize the MaxEnt objective %
by reducing the KL-divergence between $\pi$ and the canonical distribution of its current soft Q-function. This is equivalent to maximizing the expected soft Q-function's value corrected by the policy's entropy, resembling a regularized policy gradient objective \citep{dpg, ddpg}:
\begin{equation} \label{eq:policy_impr_obj}
    \argmax_\theta \E_{s, a\sim \pi_\theta(\cdot|s)}\left[Q^\pi_\phi(s,a) + \alpha H(\pi_\theta(a|s))\right].
\end{equation}
The policy is usually modeled with a neural network outputting the parameters of some tractable distribution, such as a factorized Gaussian, $\pi_\theta(\cdot|s)= N(\mu_\theta(s); \Sigma_\theta(s))$. This practice allows to efficiently approximate the gradients from Eqn.~\ref{eq:policy_impr_obj} via the reparameterization trick \citep{vae}. We consider the off-policy RL setting, where the agent alternates learning with storing new experience in a data buffer, $D$. We refer the reader to \citet{sac, sac-alg} for further derivation and practical details.

\section{Policy Gradient with serial reasoning}

\label{sec:3description}

\subsection{Reasoning as a Markov chain}
\label{subsec:res-as-MC}

We introduce \textit{Serial Markov Chain Reasoning}, a new framework to model agent behavior, based on conceptualizing action selection as an adaptive, sequential process which we refer to as \textit{reasoning}. Instead of using a traditional policy, the agent selects which action to execute by maintaining an internal \textit{action-belief} and a \textit{belief transition (BT-) policy}, $\pi^{b}(a'|a, s)$. During the \textit{reasoning process}, the agent updates its action-belief for a series of \textit{reasoning steps} by sampling a new action with the BT-policy $\pi^b$ taking both environment state and previous action-belief as input. %
We naturally represent this process with a \textit{reasoning Markov chain} (RMC), a discrete-time Markov chain over different action-beliefs, with transition dynamics given by the BT-policy. %
Hence, for any input environment state $s$ and initial action-belief $a_0$, the n-step transition probabilities of the RMC for future \textit{reasoning steps} $n=1, 2, 3, ...$ are defined as:
\begin{equation}
        \pi^b_n(a|a_0, s) = \int_A \pi^{b}(a|a', s) \pi^b_{n-1}(a'|a_0, s) da', \quad \textit{ for } n> 1\textit{, and } \quad \pi^b_{1}=\pi^b.
\end{equation}
Given a compact action space and a BT-policy with a non-zero infimum density, we can ensure that as the number of reasoning steps grows, the probability of any action-belief in the RMC \textit{converges} to some \textit{steady-state probability} which is independent of the initial action-belief.\footnote{This is unrelated to the \textit{steady-state} distribution for infinite-horizon MDPs considered in prior work \citep{suttonbarto}.} We denote this \textit{implicit} probability distribution as the \textit{steady-state (SS-) policy}, symbolized by $\pi^{s}(a|s)$: %

\begin{lemma} %
\label{lemma:steady_state}
\textbf{Steady-state convergence.} For any environment state $s$, consider a reasoning Markov chain (RMC) defined on a compact action space $A$ with transition probabilities given by $\pi^{b}(a'|a, s)$. Suppose that $\inf \{\pi^{b}(a'|a, s): a', a\in A\} > 0$. Then there exists a steady-state probability distribution function $\pi^{s}(\cdot|s)$ such that:
\begin{equation}\label{Eq.nstepBT}
    \lim_{n\rightarrow \infty} \pi^b_n(a|a_0, s)\rightarrow \pi^{s}(a|s) \quad \textit{for all }a\in A.
\end{equation}
\begin{proof}
See Appendix~\ref{app:proofs}.
\end{proof}
\end{lemma}

\begin{figure}[t]
    \centering
    \includegraphics[width=0.99\linewidth]{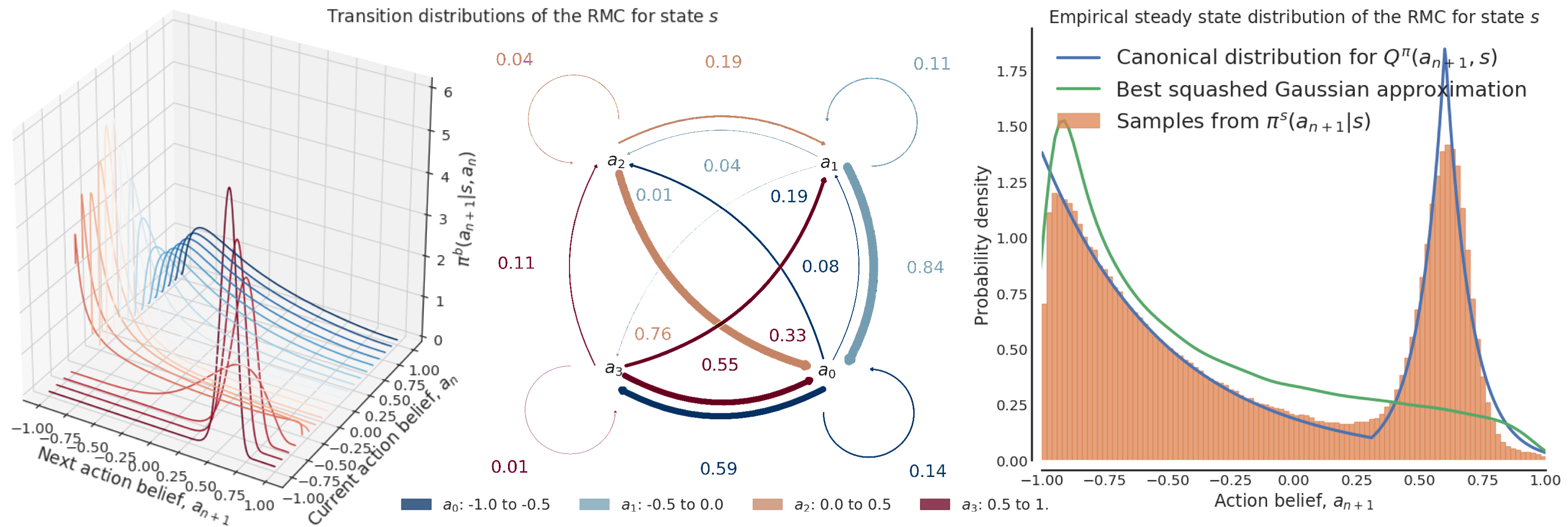} 
    \caption{\small{{(\textbf{Left}) BT-policy transition probabilities and quantized RMC state diagram. (\textbf{Right}) Sample approximation of related steady-state distribution as compared to canonical distribution of the soft Q-function.}}}
    \label{fig:bt_ss_pol}
\end{figure}%

The RMC's steady-state probabilities can be interpreted as representing the distribution of agent's behavior after an \textit{appropriate} number of reasoning steps are performed. %
In this work, we strive to optimize the agent's behavior following the MaxEnt RL framework described in Section~\ref{sec:background}. In particular, we consider learning a parameterized BT-policy, $\pi_\theta^b$, to produce appropriate transition probabilities for each environment state such that the SS-policy, $\pi_\theta^s$, from the resulting RMC optimizes:
\begin{equation} \label{eq:pg_bt_obj}
    \begin{split}
        \argmax_{\theta} J(\theta) %
        &= \E_{s, a\sim \pi^s_\theta(\cdot|s)}\left[Q_\phi^{s}(s,a) + \alpha H(\pi^s_{\theta}(a|s))\right].
    \end{split}
\end{equation}
Here, $Q_\phi^{s}$ is a parameterized soft Q-function for the agent's behavior from $\pi^s$, which we learn by minimizing a squared \textit{soft} Bellman loss utilizing delayed parameters $\phi'$ and samples from $\pi_\theta^s$:
\begin{equation}\label{eq:q_fn_bt_obj}
\small
\begin{split}
    \argmin_\phi J(\phi)= \E_{s, a, s'}\left[(Q_\phi^{s}(s, a) -
    \left(r(s, a) + \gamma\E_{a'\sim \pi^s_\theta(\cdot|s)}\left[Q_{\phi'}^{s}(s', a') +\alpha H(\pi^s_{\theta}(a'|s')\right]\right)\right]^2.\\
\end{split}
\end{equation}
In Fig.~\ref{fig:bt_ss_pol}, we illustrate the relationship between a learned BT-policy, the corresponding SS-policy, and the soft Q-function in a 1-dimensional toy task (see App.~\ref{app:sspg_impl} for details). In this example, the BT-policy is parameterized as a simple squashed Gaussian distribution, with unimodal transitions between consecutive action beliefs (Fig.~\ref{fig:bt_ss_pol}, Left). We obtain samples of agent behavior (the SS-policy) by performing a series of reasoning steps, using the BT-policy to simulate the RMC until we approach steady-state convergence. By plotting the resulting empirical distribution of agent behavior, we see it closely matches the multi-modal, non-Gaussian canonical distribution from its soft Q-function (Fig.~\ref{fig:bt_ss_pol}, Right). This example shows how the expressive power of agent behavior in our framework can go far beyond the BT-policy's simple parameterization, enabling for the effective maximization of complex and multi-modal MaxEnt objectives. %

\subsection{Learning the belief transition policy}

\label{subsec:learning_bt_pol}

We propose a new method to estimate the policy gradient of the BT-policy, $\pi^{b}_\theta$, for optimizing the steady-state MaxEnt objective described in Section~\ref{subsec:res-as-MC}. %
We note that the gradient from Eq.~\ref{eq:pg_bt_obj} involves differentiating through an expectation of the steady-state policy, $\pi^{s}_\theta$. However, $\pi^{s}_\theta$ is only \textit{implicitly} defined, and its connection with the actual BT-policy or its parameters does not have a tractable closed-form expression. To approach this problem, we introduce a family of \textit{n-step extensions} to the soft Q-function, $Q_n^s:S\times A\mapsto \R$ for $n=0,1,2,\dots$, defined as:
\begin{gather} \label{Eq.Qidefn}
Q^s_n(s, a) = \int_A \pi_n^b(a'|a,s) Q^s_\phi(s,a') da', \quad %
\textit{with} \quad\g Q^s_n(s, a) = \mathbf{0}.%
\end{gather}
Intuitively, each \textit{n-step soft Q-function} $Q^s_n(s, a)$ outputs the \textit{expected soft Q-value} after performing $n$ reasoning steps in the RMC from the initial action-belief $a$. However, we treat the output of each n-step soft Q-function as being \textit{independent} of the actual parameters of the BT-policy, $\theta$. Hence, we can interpret computing  $Q^s_n(s, a)$ as simulating the RMC with a \textit{fixed and immutable copy} of the current $\pi^b_\theta$. We use this definition to provide a convenient notation in the following new Theorem that expresses the policy gradient without differentiating through $\pi_\theta^{s}$:
\begin{theorem} \label{thm:BT-POLGRAD}
\textbf{Steady-state policy gradient.} Let $\pi_\theta^b(\cdot|a, s)$ be a parameterized belief transition policy which defines a reasoning Markov chain with a stationary distribution given by the steady-state policy $\pi_\theta^s(\cdot|s)$. Let $Q^s$ be a real function defined on $S\times A$, with a family of n-step extensions $\{Q^s_n\}$ as defined in Eq.~\ref{Eq.Qidefn}. Suppose $\pi^b$, $Q^s$ and their gradient with respect to $\theta$ (denoted $\g$) are continuous and bounded functions. Then
\begin{equation}
\begin{split}\label{Eq.thm_eq}
  \g \E_{a\sim \pi_\theta^s(\cdot|s)}\left[Q^s(s, a)\right] = \E_{a\sim \pi_\theta^s(\cdot|s)}  \left[\lim_{N\to \infty} \sum_{n=0}^{N} \g \E_{a'\sim \pi^b_\theta(\cdot|a, s)} \left[Q^s_n(s, a')\right]\right]. \\
\end{split}
\end{equation}
\begin{proof}
See Appendix~\ref{app:proofs}.
\end{proof}
\end{theorem}
Using Lemma~\ref{lemma:steady_state} (\textit{steady-state convergence}), we can approximate the policy gradient expression in Eq.~\ref{Eq.thm_eq} with an arbitrarily small expected error using a finite number of n-step soft Q-functions, i.e., $N$ (see App.~\ref{app:proofs}). An intuition for this property follows from the fact that for large enough $n$, Lemma~\ref{lemma:steady_state} implies that $\pi^b_n(a|a', s)\approx \pi^{s}_\theta(a|s)$ and, thus, $Q_n^{s}(s, a')\approx \int_A \pi^b_\theta{s}(a|s) Q^s_\phi(s,a) da$. Therefore, the value of each $Q_n^{s}(s, a')$ will be independent of the BT-policy's action $a'$, such that $\g \E_{a'\sim \pi^b_\theta(\cdot|a, s)} \left[Q_n^s(s, a')\right]\approx \mathbf{0}$. %
In other words, each subsequent step in the RMC introduces additional randomness that is independent of $a'$, causing a warranted \textit{vanishing gradient} phenomenon \citep{bengio-vanishing-grad} which culminates with converging to $\pi^s_\theta$. Using a similar notation as \citet{sac-alg}, we apply the reparameterization trick \citep{vae} to express the BT-policy in terms of a deterministic function $f_\theta^b(a, s, \epsilon)$, taking as input a Gaussian noise vector $\epsilon$. This allows to rewrite the gradient in each inner expectation in the sum from Eq.~\ref{Eq.thm_eq} as:
\begin{equation}\label{Eq.reparam_og}
    \g \E_{a'\sim \pi^b_\theta(\cdot|a, s)} \left[Q^s_n(s, a')\right] = \E_{\epsilon_0\sim N(0,1)} \left[\nabla_{a_0}Q^s_n(s, a_0) \g f_\theta^b(a, s, \epsilon_0)\right], %
\end{equation}
where $a_0=f_\theta^b(a, s, \epsilon _0)$. We can apply the same reparameterization for all n-step soft Q-functions, to establish a new relationship between the gradient terms $\nabla_{a_0} Q^s_n(s, a_0)$: %
\begin{align} \label{Eq.rec_rel_g}
    \nonumber
    \nabla_{a_0} Q^s_n(s, a_0) &= \nabla_{a_0} \int_A \pi_{n}^b(a_n|a_0,s) Q_\phi^s(s,a_n) da_n 
    = \nabla_{a_0} \int_A \pi^b(a_1| a_0, s)  Q^s_{n-1}(s,a_1) da_1 \\
    &= \E_{\epsilon_1} \left[\nabla_{a_1} Q^s_{n-1}(s,a_1) \nabla_{a_0} f^b(a_0, s, \epsilon_1) \right], \quad \textit{where, } a_1=f(a_0, s, \epsilon_1).
\end{align}
In Eq.~\ref{Eq.rec_rel_g}, we purposefully omit the dependence of $f^b$ and $\pi^b$ from $\theta$ since each $Q^s_n$ term is a \textit{local} approximation of the RMC that does not depend on $\theta$ (as defined in Eq.~\ref{Eq.Qidefn}). By recursively applying this relationship (Eq.~\ref{Eq.rec_rel_g}) to $\nabla_{a_1} Q^s_{n-1}(s,a_1)$ and all subsequent gradient terms we obtain:
\begin{align} \label{Eq.grad_prod}
    \nabla_{a_0} Q^s_n(s, a_0) = \E_{\epsilon_1, \dots, \epsilon_n} \left[\nabla_{a_n} Q^s_{\phi}(s,a_n) \prod^{n-1}_{i=0} \nabla_{a_i} f^b(a_{i}, s, \epsilon_{i+1})\right], %
\end{align}
where $a_i=f^b(a_{i-1}, s, \epsilon_{i})$ for $i=1,\dots,n$. By combining Eq.~\ref{Eq.reparam_og} and Eq.~\ref{Eq.grad_prod}, we can thus reparameterize and express the whole sum in Eq.~\ref{Eq.thm_eq} as:
\begin{equation}
\label{Eq.whole_reparam}
\begin{gathered}
    \g \E_{a\sim \pi_\theta^s(\cdot|s)}\left[Q^s(s, a)\right] \approx 
    \sum_{n=0}^{N} \g \E_{a'\sim \pi^b_\theta(\cdot|a, s)} \left[Q^s_n(s, a')\right] \\
    =\E_{\epsilon_0,\dots,\epsilon_N}\left[\sum^N_{n=0} \nabla_{a_n} Q^s_{\phi}(s,a_n) \left(\prod^{n-1}_{i=0} \nabla_{a_i} f^b(a_{i}, s, \epsilon_{i+1})\right)\g f_\theta^b(a, s, \epsilon_0)\right].
\end{gathered}
\end{equation}
Eq.~\ref{Eq.whole_reparam} intuitively corresponds to differentiating through each $Q^s_n(s, a')$ term by \textit{reparameterizing the RMC}. Hence, to get a sample estimate of the policy gradient we can simulate the reparameterized RMC for $N$ reasoning steps to obtain $a_1, ..., a_N$, compute each $Q^s_\phi(s, a_n)$ term, and backpropagate (e.g., with \textit{autodifferentiation}). Following \citet{sac, sac-alg}, we can apply Theorem~\ref{thm:BT-POLGRAD} and easily extend the same methodology to estimate the MaxEnt policy gradient from Eq.~\ref{eq:pg_bt_obj} that also involves an extra entropy term. We include this alternative derivation in App.~\ref{app:proofs} for completeness.

\subsection{Action selection and temporal consistency}

\label{subsec:action_sel_temp_cons}
To collect experience in the environment, we propose to perform reasoning with the BT-policy starting from \textit{a set} of different initial action-beliefs $\{a^0_0, ..., a^M_0\}$. We \textit{batch} this set as a single input matrix, $\mathbf{a_0}$, to make effective use of parallel computation. %
To reduce the required number of reasoning steps and facilitate detecting convergence to $\pi^s_\theta$, we identify two desirable properties for the distribution of action-beliefs in $\mathbf{a_0}$. In particular, initial action-beliefs should 1) be likely under $\pi^s_\theta$, and 2) cover diverse modes of $\pi^s_\theta$. Property (1) should logically accelerate reasoning by providing the BT-policy with already-useful information about optimal behavior. %
Property (2) %
serves to provide the BT-policy with initial information of diverse behavior, which facilitates convergence detection (Sec.~\ref{subsec:ss_conv_gelrub}) and expedites reasoning even if the RMC has slow mixing times between multiple modes. To satisfy these properties, we use a simple effective heuristic based on common temporal-consistency properties of MDPs \citep{action-reps-policy-factor, epsgreedytemp}. Especially in continuous environments, actions tend to have small individual effects, making them likely relevant also for environment states experienced in the near future. Thus, we propose storing past action-beliefs in a fixed sized buffer, called the \textit{short-term action memory}, $\hat{A}$, and use them to construct $\mathbf{a_0}$. We find this strategy allows to effectively regulate the initial action-beliefs quality and diversity through the size of $\hat{A}$, accelerating convergence at negligible cost. %

\subsection{Detecting convergence to the steady-state policy}
\label{subsec:ss_conv_gelrub}

\begin{figure}
\begin{minipage}[t]{0.46\textwidth}
\vspace{0pt}
\begin{algorithm}[H]
    \footnotesize
    \caption{Agent Acting}
    \label{alg:act_sel}
    \begin{algorithmic}
    \State \textbf{input:} $s$, current state%
    \State $\mathbf{a_0} \sim \hat{A}$
    \State $N\gets 0$
    \State $R^p\gets+\infty$
    \While{$R^p>1.1$}
    \State $\mathbf{a_{N+1}}\sim \pi^b_\theta(\cdot| \mathbf{a_{N}})$
    \State $N\gets N+1$
    \State Update $R^p$ with $\mathbf{a_{1:N}}$\Comment{Eq.\ref{Eq.psrf}}
    \EndWhile
    \State $\hat{N}\gets \rho\hat{N}+(1-\rho)N$ \Comment{$\rho \in[0, 1)$}
    
    \State $\hat{A} \gets \hat{A} \cup \mathbf{a_{1:N}}$
    \State \textbf{output:} $a \sim \mathbf{a_{1:N}}$
    \end{algorithmic}
\end{algorithm}
\end{minipage}
\hfill
\begin{minipage}[t]{0.50\textwidth}
\vspace{0pt}
\begin{algorithm}[H]
    \footnotesize
    \caption{Agent Learning}
    \label{alg:learn}
    \begin{algorithmic}
    \State \textbf{input:} $D$, data buffer %
    \State $(s, a, s', r) \sim D$
    \State $a_0 \sim \pi_\theta^b(\cdot|a, s')$
    \For{$n\gets 0, \ceil{\hat{N}}$ }
        
        \State $Q_n^s\gets Q^s_\phi(s', a_n)$ \Comment{Eq.~\ref{Eq.Qidefn}}
        \State $\epsilon_{n+1} \sim N(0,1), \quad a_{n+1} = f^b(a_n, s, \epsilon_{n+1})$
    \EndFor
    
    \State $\nabla_\theta Q^{s}_\phi \gets %
    \g (\sum^{\ceil{N}}_{n=0} Q_n^s)$ \Comment{Thm.~\ref{thm:BT-POLGRAD}}
    \State $\argmin_\theta J(\theta)$ \Comment{Eq.~\ref{eq:pg_bt_obj}}
    \State $a'\sim a_{1:\ceil{\hat{N}}}$
    \State $\argmin_\phi J(\phi)$ \Comment{Eq.~\ref{eq:q_fn_bt_obj}}
    \end{algorithmic}
\end{algorithm}
\end{minipage}
\end{figure}

A key requirement for learning and acting with BT-policies, as described in Sections~\ref{subsec:learning_bt_pol} and \ref{subsec:action_sel_temp_cons}, is the ability to determine a \textit{sufficient} number of reasoning steps ($N$) for the action-belief distribution to converge. Given the properties of the RMC, there exist different analytical methods that provide a priori bounds on the rate of convergence \citep{small-sets-conv-prop, conv-bounds-1, conv-bounds-2}. However, using any fixed $N$ would be extremely limiting as we expect the BT-policy and the properties of its resulting RMCs to continuously evolve during training. Moreover, different tasks, states, and initial action-beliefs might affect the number of reasoning steps required for convergence due to different levels of complexity for the relative decision-making problems. To account for similar conditions, in the Markov Chain Monte Carlo literature, the predominant approach is to perform a statistical analysis of the properties of the simulated chain, choosing from several established convergence diagnostic tools \citep{conv-diagnostics, conv-diagnostics-2, conv-diagnostics-3}.
Hence, we propose to employ a similar \textit{adaptive strategy} by analyzing the history of the simulated RMC to determine the appropriate number of reasoning steps. Since we apply $\pi^b_\theta$ from a diverse set of initial action beliefs  (see Section~\ref{subsec:action_sel_temp_cons}), we base our convergence-detection strategy on the seminal Gelman-Rubin (GR) diagnostic \citep{gelmanrubin} and its multivariate extension \citep{gelmanrubin-gen-mult}. In particular, the multivariate GR diagnostic computes the \textit{pseudo scale reduction factor} (PSRF), a score representing whether the statistics of a multivariate variable of interest have converged to the steady-state distribution. The intuition behind this diagnostic is to compare two different estimators of the covariance for the unknown steady-state distribution, making use of either the samples \textit{within} each individual chain and \textit{between} all different chains. Thus, as the individual chains approach the true steady-state distribution, the two estimates should expectedly get closer to each other. The PSRF measures this precise similarity based on the largest eigenvalue of their matrix product. 

For our use-case, we employ the PSRF to determine the convergence of the set of action-beliefs $\mathbf{a_{1:N}}$, as we perform consecutive reasoning steps with $\pi^b_\theta$. Following \citep{gelmanrubin-gen-mult}, we calculate the average sample covariance of the action-beliefs \textit{within} each of the parallel chains ($W$) computed from a batched set of initial action-beliefs %
$\mathbf{a_0}=[a^1_0, a^2_0, \dots a^M_0]$:
\begin{equation}
    \label{Eq.within_sample_cov}
    \bar{a}^m = \frac{1}{N}\sum^N_{n=1} a_n^m, \quad W_m = \frac{1}{N-1}\sum_{n=1}^N (a - \bar{a}^m)(a-\bar{a}^m)^T, \quad W = \frac{1}{M}\sum^M_{m=1} W_m.
\end{equation}
We compare $W$ with an unbiased estimate of the target covariance, constructed from the sample covariance \textit{between} the different parallel chains ($B$):
\begin{equation}
    \label{Eq.between_sample_cov}
    \bar{a} = \frac{1}{N\times M}\sum^N_{n=1}\sum^M_{n=1} a_n^m, \quad B = \frac{1}{M-1}\sum_{n=1}^N (\bar{a}^m - \bar{a})(\bar{a}^m - \bar{a})^T.
\end{equation}
The PSRF for $\mathbf{a_{1:N}}$ is then computed from the largest eigenvalue ($\lambda_{max}$) of the product $W^{-1}B$, as: %
\begin{equation}
    \label{Eq.psrf}
    R^p = \sqrt{\frac{N-1}{N} + \lambda_{max}(W^{-1}B)}.
\end{equation}
Thus, as the individual chains approach the distribution of $\pi^s_\theta$, the PSRF ($R^p$) will approach 1. Following \citet{gelmanrubin-gen-mult}, we use $R^p < 1.1$ as an effective criterion for determining the convergence of $\mathbf{a_{1:N}}$. In practice, we also keep a \textit{running mean} of the current number of reasoning steps for convergence, $\hat{N}$. We use $\ceil{\hat{N}}$ as the number of reasoning steps to simulate the RMC with $\pi^b_\theta$ when computing gradients from Eqs.~\ref{eq:pg_bt_obj}-\ref{eq:q_fn_bt_obj}. $\ceil{\hat{N}}$ is a \textit{safe} choice to ensure near unbiased optimization since $R^p < 1.1$ is considered a very conservative criterion \citep{gelmanrubin-revisited} and we can learn by simulating the RMC from recent actions stored in the data buffer, which are already likely close to optimal. %
We provide further details regarding our implementation and its rationale in App.~\ref{app:gel_rub_impl}. We provide a simplified summary of our adaptive reasoning process for acting and learning in Algs.~\ref{alg:act_sel}-\ref{alg:learn}.

\subsection{Advantages of serial Markov chain reasoning}
\label{subsec:smc_adv}

Based on the above specification, we identify three main conceptual advantages of our serial Markov chain reasoning framework. \textbf{1. Unlimited expressiveness.} The distribution of agent behavior given by the SS-policy $\pi^s_\theta$, is a mixture model with potentially infinitely many components. Thus, even a simple Gaussian parameterization of the BT-policy $\pi^b_\theta$ would make $\pi^s_\theta$ a \textit{universal approximator of densities}, providing unlimited expressive power to the agent \citep{FiniteMixStatAna, goodfellow-deep-learning}. \textbf{2. Adaptive computation.} The number of reasoning steps performed to reach approximate convergence is determined by the properties of each environment state's RMC. Hence, the agent can flexibly spend different amounts of computation time based on the complexity of each action-selection problem, with potential gains in both precision and efficiency. \textbf{3. Information reuse.} By storing past solutions to similar RMCs, we can initialize the reasoning process with initial action-beliefs that are already close to $\pi^s_\theta$. This allows using the temporal-consistency properties of the MDP to exploit traditionally discarded information and accelerate agent reasoning. We provide empirical validation for these properties in Section~\ref{subsec:4prop_exps}.

\section{Experimentation}

\label{sec:4exps}

\subsection{Performance evaluation}
\label{subsec:4perf_exps}

\begin{figure}[t]
    \centering
    \includegraphics[width=\linewidth]{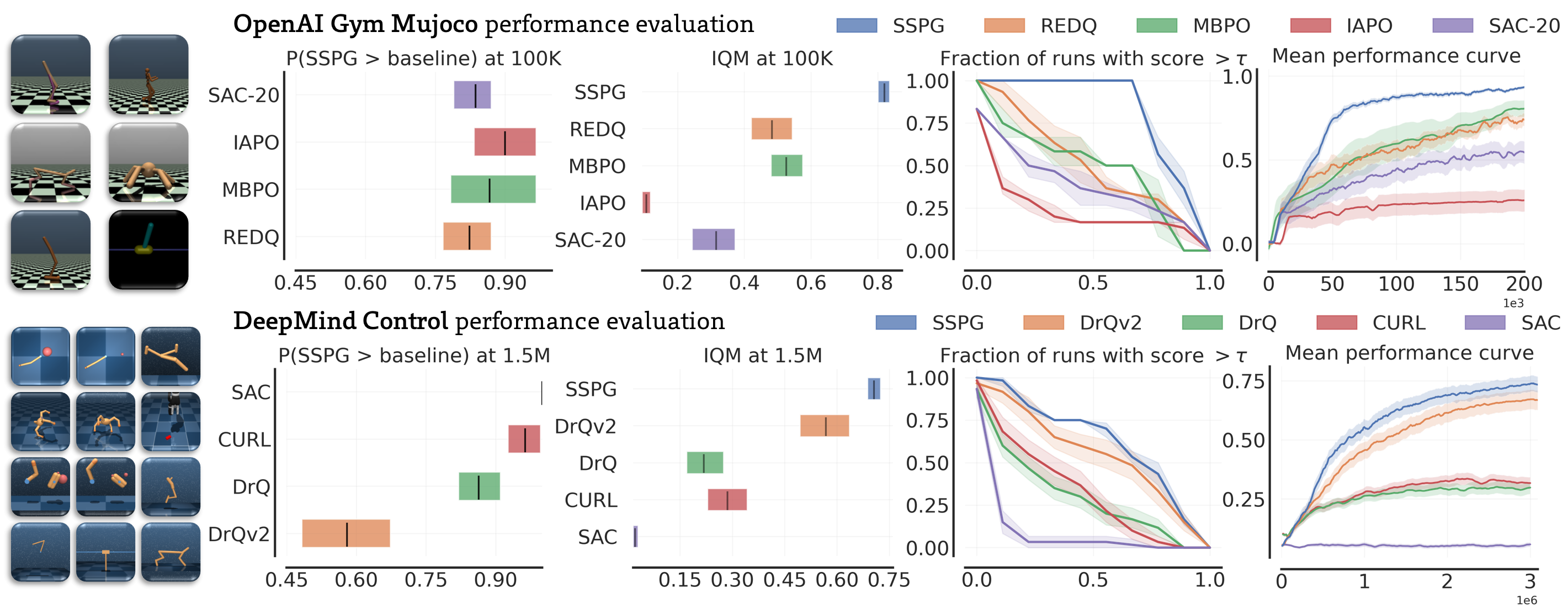} 
    \caption{\small{{Performance evaluation of SSPG and recent state-of-the-art baselines using \textit{Rliable}~\citep{precipice-rliable}. We consider six OpenAI Gym Mujoco tasks \citep{gym} (\textbf{Top})} and twelve DeepMind Control tasks from pixels \citep{dmc}  (\textbf{Bottom}).}}
    \label{fig:perf_comp}
\end{figure}%

We evaluate the \textit{serial Markov chain reasoning} framework by comparing its performance with current state-of-the-art baselines based on traditional RL. We consider 6 challenging Mujoco tasks from Gym \citep{mujoco, gym} and 12 tasks pixel-based tasks from the DeepMind Control Suite (DMC) \citep{dmc}. In both settings, we base our implementation on MaxEnt RL, replacing the traditional policy with a Gaussian BT-policy optimized with the training procedures specified in Sec.~\ref{sec:3description}. Other orthogonal design choices (e.g., network architectures) follow contemporary RL practices, we refer to App.~\ref{app:sspg_impl} or the code for full details. We call the resulting algorithm \textit{Steady-State Policy Gradient} (SSPG). 

We report the mean performance curves and aggregate metrics using the statistical tools from \textit{Rliable} \citep{precipice-rliable}. In particular, we compare normalized \textit{performance profiles} \citep{perf-profile}, \textit{interquantile  mean} (IQM), and \textit{probability of improvements} over baselines with the Mann-Whitney U statistic \citep{mannwhitneyUstat}. The reported ranges/shaded regions represent 95\% \textit{stratified bootstrap confidence intervals} (CIs) \citep{bootstrap-cis}. In App.~\ref{app:ext_res}, we provide per-task results and further statistical analysis. For each experiment, we collect the returns of SSPG over five seeds, by performing 100 evaluation rollouts during the last 5\% of steps.

\textbf{Mujoco suite.} We evaluate on a challenging set of Mujoco tasks popular in recent literature. We compare SSPG with recent RL algorithms achieving state-of-the-art sample-efficiency performance on these tasks, which utilize large critic ensembles and high update-to-data (UTD) ratios. We consider \textit{REDQ} \citep{redq} and \textit{MBPO} \citep{mbpo} for state-of-the-art algorithms based on the traditional model-free and model-based RL frameworks. We also compare with \textit{iterative amortized policy optimization} (IAPO) \citep{iterativeAmortizedPolOptim}, in which the agent performs iterative amortization to optimize its policy distribution \citep{iterativeAmort}. This procedure for action selection is more computationally involved than our agent's \textit{reasoning process}, as it requires both evaluating the policy and computing gradients at several iterations. Yet, as IAPO is still based on the traditional policy gradient framework, its benefits are solely due to reducing the \textit{amortization gap} with an alternative action inference procedure. To ground different results, we also show the performance of the seminal \textit{Soft Actor-Critic} (SAC) algorithm \citep{sac-alg}, upon which all considered policy gradient baselines are based on. To account for the additional computational cost of training an agent with serial Markov chain reasoning, we use a UTD ratio that is \textit{half} the other algorithms. On our hardware, this makes SSPG faster than all other modern baselines (see App.~\ref{app:ext_res}).

Figure~\ref{fig:perf_comp} (Top) shows the performance results after 100K environment steps. Individual scores are normalized using the performance of SAC after 3M steps, enough to reach convergence in most tasks. SSPG considerably outperforms all prior algorithms with \textit{statistically meaningful} gains, as per the conservative Neyman-Pearson statistical testing criterion \citep{stat-significance-neyman-pearson}. Furthermore, SSPG even \textit{stochastically dominates} all considered state-of-the-art baselines \citep{stoch-dom}. We obtain similar results evaluating at 50K and 200K steps (App.~\ref{app:ext_res}). In comparison, IAPO obtains lower performance than other non-iterative baselines while being the most compute-intensive algorithm. This indicates that, for sample-efficiency, only reducing the amortization gap beyond direct estimation might not provide significant benefits. Instead, serial Markov chain reasoning's improved expressivity and flexibility appear to considerably accelerate learning, yielding state-of-the-art performance in complex tasks.

\textbf{DeepMind Control suite.} To validate the generality of our framework, we also evaluate on a considerably different set of problems: 12 \textit{pixel-based} DMC tasks. We follow the recent task specifications and evaluation protocols introduced by \citet{drqv2}. We compare SSPG with \textit{DrQv2} \citep{drqv2}, the current state-of-the-art policy gradient algorithm on this benchmark, which employs a deterministic actor and hand-tuned exploration. We also compare with additional baselines that, like SSPG, are based on MaxEnt RL: \textit{DrQ} \citep{drq}, \textit{CURL} \citep{curl}, and a convolutional version of \textit{SAC} \citep{sac-alg}. 

Figure~\ref{fig:perf_comp} (Bottom) shows the performance results after 1.5M environment steps. DMC tasks yield returns scaled within a set range, $[0, 1000]$, which we use for normalization. Remarkably, also in this domain, SSPG attains state-of-the-art performance with statistically significant improvements over all baselines. Unlike for the Mujoco tasks, the other considered algorithms based on MaxEnt RL underperform as compared to the deterministic DrQv2, a result \citet{drqv2} attributed to ineffective exploration. In contrast, SSPG yields performance gains \textit{especially} on sparser reward tasks where the other baselines struggle (see App.~\ref{app:ext_res}). These results validate the scalability of our framework to high-dimensional inputs and its ability to successfully complement MaxEnt RL.

\subsection{Properties of serial Markov chain reasoning}

\begin{figure}[t]
    \centering
    \includegraphics[width=\linewidth]{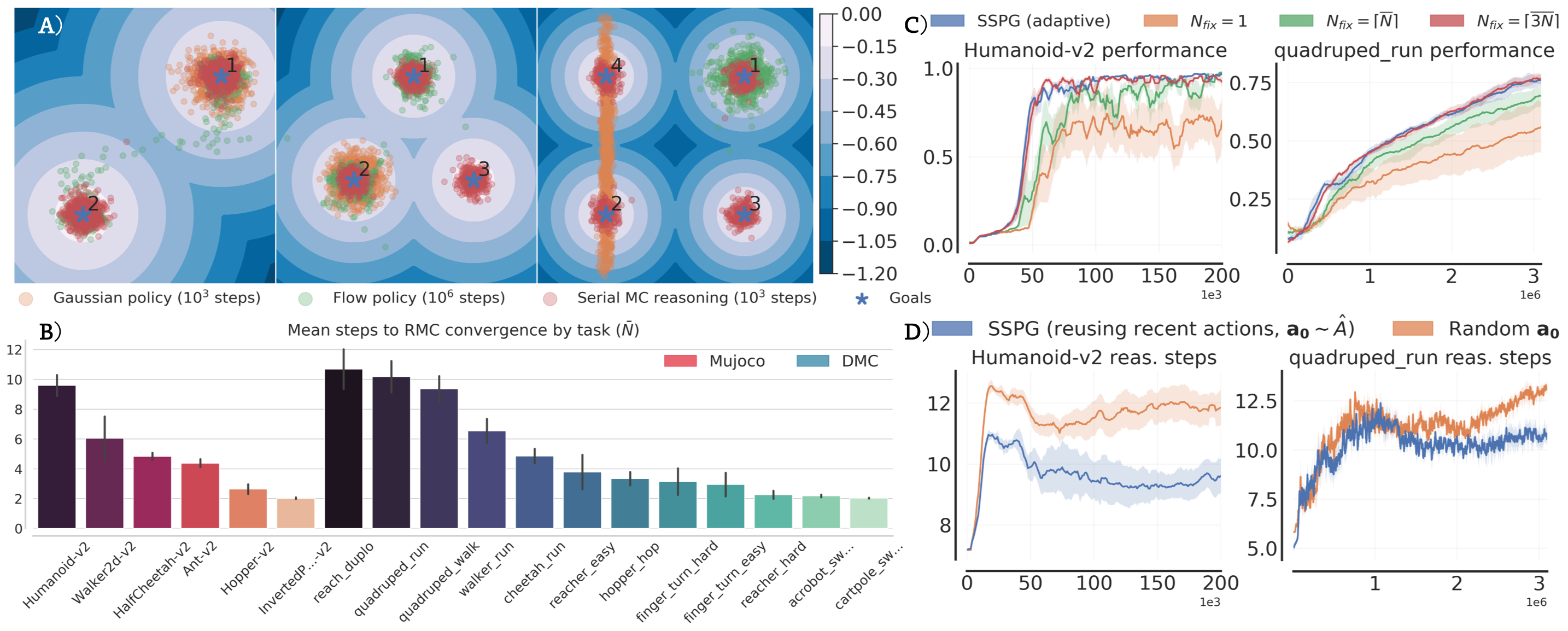} 
    \caption{\small{(\textbf{A}) Samples visualizations for learned policies in \textit{positional bandits} of increasing complexity. (\textbf{B}) Mean number of \textit{reasoning steps} required to reach convergence in each task with SSPG. (\textbf{C}) Mean performance ablating the adaptive strategy for detecting \textit{reasoning convergence} and using fixed numbers of steps. (\textbf{D}) Number of \textit{reasoning steps} throughout training with and without reusing recent actions as initial \textit{action-beliefs}.}}
    \label{fig:ana_comp}
\end{figure}%

\label{subsec:4prop_exps}

We test if theorized benefits of our framework (Sec.~\ref{subsec:smc_adv}) hold in practical settings with deep networks and stochastic optimization. We provide further ablation studies and analysis of SSPG in App.~\ref{app:abl}.

\textbf{1. Policy expressiveness.} First, we test the expressiveness of the behavior learned with SSPG using a Gaussian BT-policy. We design a series of single-step toy RL problems where the agent needs to position itself on a small 2D environment with a reward function based on unknown goal locations, which we name \textit{positional bandits} (see App.~\ref{app:sspg_impl} for details). The objective of these experiments is to isolate how our framework compares with traditional policies for MaxEnt RL to explore the environments and learn to match the true canonical distributions of returns. As displayed in Fig.~\ref{fig:ana_comp} A, even in highly multi-modal positional bandits, the SS-policy successfully learns to visit all relevant goals with similar frequencies. Furthermore, quantizing the state space around the goals reveals that the relative RMC intuitively learns to \textit{transition between action-beliefs that visit the different goals} as reasoning progresses, with a transition matrix matching a \textit{cyclic permutation} (App.~\ref{app:add_ana}). In comparison, a squashed Gaussian policy expectedly fails to capture the complexity of the canonical distribution, with samples either collapsing to a single mode or covering large suboptimal parts of the action space. We also show results for a policy based on normalizing flows \citep{normflowsSEM, realnvp}, modeled with a deep expressive network (App.~\ref{app:sspg_impl}). After several attempts, we find these models require orders of magnitude more training iterations and data to learn any behavior that is more complex than a uni-modal distribution. Yet, even after increasing training by a factor of 1000, we still observe the flow policy distribution collapsing in the more complex positional bandits. We attribute our findings to training inefficiencies from a lack of proper inductive biases for flow models in the \textit{non i.i.d.} RL problem setting \citep{flows-review-ind-biases}. In particular, as flows can assign arbitrarily low probability mass to some regions of the action space, initial local optima can greatly hinder future exploration, exacerbating coverage of the data buffer distribution in a vicious circle.

\textbf{2. Policy adaptivity.} Second, we examine the adaptivity of our framework for tackling decision-making problems with different complexities. We compare the average number of reasoning steps ($\bar{N}$) performed by SSPG for each task from Sec.~\ref{subsec:4perf_exps} (Fig.~\ref{fig:ana_comp} B). We identify a general correlation between task difficulty and reasoning computation, with complex robotic manipulation and humanoid locomotion problems requiring the most steps. By concentrating on two representative tasks, we validate the effectiveness of the reasoning process and our adaptive convergence detection strategy with an ablation study where we train SSPG using a \textit{fixed} number of reasoning steps $N_{fix}\in \{1, \ceil{\bar{N}}, \ceil{3\bar{N}}\}$. For the case $N_{fix}=1$, which closely resembles traditional RL, we use double the UTD ratio to improve performance and offset any training-time gains from multi-step reasoning. As shown in Fig.~\ref{fig:ana_comp} C, increasing $N_{fix}$ yields clear performance improvements, validating that agents can greatly benefit from performing longer reasoning processes. Furthermore, our adaptive SSPG attains the same performance as $N_{fix}=\ceil{3\bar{N}}$ and visibly outperforms $N_{fix}=\ceil{\bar{N}}$. These results show how different action selection problems require different amounts of \textit{reasoning computation} and validate the practical effectiveness of our adaptive strategy to detect steady-state convergence. We obtain analogous findings for additional tasks and values of $N_{fix}$ in App.~\ref{app:add_ana}.

\textbf{3. Solution reuse.} Last, we examine the effects of the \textit{short-term action memory} buffer ($\hat{A}$) to sample initial action beliefs ($\mathbf{a_0}$) in two tasks. We evaluate ablating $\hat{A}$, randomly re-initializing $\mathbf{a_0}$ from a uniform distribution. While there are only minor differences performance-wise between the two approaches (App.~\ref{app:add_ana}), sampling $\mathbf{a_0}$ from the short-term action memory considerably decreases the number of reasoning steps for convergence (Fig.~\ref{fig:ana_comp} D). Moreover, we observe the gap in reasoning efficiency expands throughout training as the agent's steady-state behavior further improves for the target task. This result validates that a simple temporal heuristic can provide considerable efficiency benefits, amortizing the additional computational cost of our powerful new framework.

\section{Related work}

\label{sec:related}

There have been several prior attempts to extend ubiquitous Gaussian policies \citep{mpo, sac, sac-alg, slac} with simple normalizing flows \citep{normflowsSEM, realnvp}, both to improve expressiveness \citep{TRPO-norm-flows, PG-norm-flows} and to instantiate behavior hierarchies \citep{latent-norm-flows-hier}. Yet, the expressiveness of normalizing flows is coupled with some training challenges \citep{flows-review-ind-biases}, which we show can lead to premature convergence to suboptimal solutions in RL (Sec.~\ref{subsec:4prop_exps}). Other works also considered entirely replacing policy models with gradient-free \citep{qt-opt} or gradient-based optimization over the predicted values \citep{rl-energy-based}. \citet{iterativeAmortizedPolOptim} similarly considered \textit{learning} an optimizer to \textit{infer} Gaussian behavior \citep{rlasinf6-levineTut} with iterative amortization \citep{iterativeAmort}. However, while all these works consider alternative modeling of agent behavior, they are still based on the traditional RL framework of representing decision-making as the output of a \textit{fixed} process. Instead, our work entails a conceptually different approach and enables implicit modeling of agent behavior as the result of an \textit{adaptive} reasoning process, orthogonally providing agents also with additional flexibility to scale computation based on the properties of each individual input state.

Outside RL, there have been efforts to model generation processes with parameterized Markov chains learned to revert fixed noise injection processes acting on data  \citep{outRLrel0-DenAE, outRLrel1-GSN, outRLrel2-INFUSION, outRLrel3-VARWALKBACK, outRLrel4-SHORTHMC}. Based on this framework, diffusion models \citep{diffSEM0, diffREL_scoreMATCHermon, diff1DDPM} recently achieved remarkable results for image generation \citep{diffSUCC0, diff1DDPM}. While applied to inherently different problem settings, these works share some conceptual resemblances with our framework and highlight the vast \textit{scaling} potential of implicit modeling.

\section{Conclusion}

\label{sec:conclusion}

We introduced \textit{serial Markov chain reasoning}, a novel framework for modeling agent behavior in RL with several benefits. We showed our framework allows an agent to 1) learn arbitrary continuous action distributions, 2) flexibly scale computation based on the complexity of individual action-selection decisions, and 3) re-use prior solutions to accelerate future reasoning. Hence, we derived \textit{SSPG} an off-policy maximum entropy RL algorithm for serial Markov chain reasoning, achieving state-of-the-art performance on two separate continuous control benchmarks. While for problems with discrete action spaces simple multinomial policy distributions already provide unlimited expressivity, we note that the inherent computational adaptivity of our framework could still yield benefits over traditional fixed policies in these settings. Furthermore, we believe our motivation and early results provide a strong argument for the future potential of serial Markov chain reasoning, even beyond off-policy RL and simulation tasks. We provide our implementation for transparency and to facilitate future extensions at \hyperlink{sites.google.com/view/serial-mcr/}{\texttt{sites.google.com/view/serial-mcr/}}.

\section*{Acknowledgments}

We thank Johannes Lutzeyer for providing valuable feedback on an earlier draft of this work. Edoardo Cetin would like to acknowledge the support from the Engineering and Physical Sciences Research Council [EP/R513064/1]. Oya Celiktutan would also like to acknowledge the support from the LISI Project, funded by the Engineering and Physical Sciences Research Council [EP/V010875/1]. Furthermore, we thank Toyota Motor Europe and Toyota Motor Corporation for providing support towards funding the utilized computational resources.

\bibliography{main}

\section*{Checklist}

\begin{enumerate}

\item For all authors...
\begin{enumerate}
  \item Do the main claims made in the abstract and introduction accurately reflect the paper's contributions and scope?
    \answerYes{See Section~\ref{sec:3description} for the derivation of our new framework and Section~\ref{sec:4exps} for empirical evaluation.}
  \item Did you describe the limitations of your work?
    \answerYes{We describe limitations related to our implementation in Sections~\ref{app:gel_rub_impl} and \ref{app:sspg_impl} of the Appendix. We also discuss limitations related to computational costs in Section~\ref{app:ext_res} of the Appendix.}
  \item Did you discuss any potential negative societal impacts of your work?
    \answerYes{See Section~\ref{app:soc_imp} of the Appendix.}
  \item Have you read the ethics review guidelines and ensured that your paper conforms to them?
    \answerYes{}
\end{enumerate}

\item If you are including theoretical results...
\begin{enumerate}
  \item Did you state the full set of assumptions of all theoretical results?
    \answerYes{}
        \item Did you include complete proofs of all theoretical results?
    \answerYes{See Section~\ref{app:proofs} of the Appendix.}
\end{enumerate}

\item If you ran experiments...
\begin{enumerate}
  \item Did you include the code, data, and instructions needed to reproduce the main experimental results (either in the supplemental material or as a URL)?
    \answerYes{We include code and instructions in the supplementary material of our submission. We will provide a link to our open-sourced implementation at \hyperlink{sites.google.com/view/serial-mcr/}{\texttt{sites.google.com/view/serial-mcr/}}}.
  \item Did you specify all the training details (e.g., data splits, hyperparameters, how they were chosen)?
    \answerYes{See Section~\ref{app:sspg_impl} of the Appendix.}
        \item Did you report error bars (e.g., with respect to the random seed after running experiments multiple times)?
    \answerYes{}
        \item Did you include the total amount of compute and the type of resources used (e.g., type of GPUs, internal cluster, or cloud provider)?
    \answerYes{See Section~\ref{app:ext_res} of the Appendix.}
\end{enumerate}

\item If you are using existing assets (e.g., code, data, models) or curating/releasing new assets...
\begin{enumerate}
  \item If your work uses existing assets, did you cite the creators?
    \answerYes{}
  \item Did you mention the license of the assets?
    \answerYes{We include the licenses with the supplementary material.}
  \item Did you include any new assets either in the supplemental material or as a URL?
    \answerYes{}
  \item Did you discuss whether and how consent was obtained from people whose data you're using/curating?
    \answerNA{}{}
  \item Did you discuss whether the data you are using/curating contains personally identifiable information or offensive content?
    \answerNA{}
\end{enumerate}

\item If you used crowdsourcing or conducted research with human subjects...
\begin{enumerate}
  \item Did you include the full text of instructions given to participants and screenshots, if applicable?
    \answerNA{}
  \item Did you describe any potential participant risks, with links to Institutional Review Board (IRB) approvals, if applicable?
    \answerNA{}
  \item Did you include the estimated hourly wage paid to participants and the total amount spent on participant compensation?
    \answerNA{}
\end{enumerate}

\end{enumerate}

\newpage
\appendix

\section*{Appendix}

\section{Proofs and extensions}
\label{app:proofs}

\subsection{Lemma~\ref{lemma:steady_state}, steady-state convergence}
\label{app:subsec:app:lemma_p}
\textit{For any environment state $s$, consider a reasoning Markov chain (RMC) defined on a compact action space $A$ with transition probabilities given by $\pi^{b}(a'|a, s)$. Suppose that $\inf \{\pi^{b}(a'|a, s): a', a\in A\} > 0$. Then there exists a steady-state probability distribution function $\pi^{s}(\cdot|s)$ such that:}
\begin{equation}%
    \lim_{n\rightarrow \infty} \pi^b_n(a|a_0, s)\rightarrow \pi^{s}(a|s) \quad \textit{for all }a\in A.
\end{equation}
\begin{proof}
Let $\delta =\inf \{\pi^{b}(a'|a, s): a', a\in A\}>0$ and let $m$ be the \textit{Lebesgue measure}. Since $A$ is compact we have that $m(A)=|A|$ is finite. Thus, let $v$ be the uniform probability measure on $A$, we have that:
\begin{equation*}
    \pi^b(a'|a) \geq \delta\times|A|\times v(a) \quad \text{for all }a',a\in A.
\end{equation*}
Hence we have that the resulting RMC is ergodic, irreducible, and aperiodic. Furthermore, this shows that the \textit{entire action space} is a \textit{small set} \citep{small0, small1book} of the reasoning Markov chain. This property directly implies the existence of a unique $\pi^s$ and quantifies an exponential bound on the rate of convergence \citep{small-sets-conv-prop, small-set-mod-defn}:
\begin{equation}\label{eq:app:exp_conv_dtv}
    ||\pi_n^b(a|a_0, s) - \pi^s(a|s)||_{TV} \leq (1-\delta\times|A|)^n, \quad \text{for all } a,a_0\in A,
\end{equation}
where $||\pi_n^b(a|a_0, s) - \pi^s(a|s)||_{TV}$ represents the \textit{total variation distance} between the n-step transition probabilities of the RMC and the SS-policy. We refer to \citet{smallNUMM} for a full formal derivation. Lemma~\ref{lemma:steady_state} clearly follows from the above results.

\end{proof}

\subsection{Theorem~\ref{thm:BT-POLGRAD}, steady-state policy gradient}

\textit{Let $\pi_\theta^b(\cdot|a, s)$ be a parameterized belief transition policy which defines a reasoning Markov chain with a stationary distribution given by the steady-state policy $\pi_\theta^s(\cdot|s)$. Let $Q^s$ be a real function defined on $S\times A$, with a family of n-step extensions $\{Q^s_n\}$ as defined in Eq.~\ref{Eq.Qidefn}. Suppose $\pi^b$, $Q^s$ and their gradient with respect to the parameters $\theta$ (denoted $\g$) are continuous and bounded functions.}

\quad \textit{Then:}
\begin{equation*}
\begin{split}
  \g \E_{a\sim \pi_\theta^s(\cdot|s)}\left[Q^s(s, a)\right] = \E_{a\sim \pi_\theta^s(\cdot|s)}  \left[\lim_{N\to \infty} \sum_{n=0}^{N} \g \E_{a'\sim \pi^b_\theta(\cdot|a, s)} \left[Q^s_n(s, a')\right]\right]. \\
\end{split}
\end{equation*}

\newcommand{\bt}{\pi^b_\theta}
\renewcommand{\ss}{\pi^s_\theta}
\newcommand{\qn}{Q^s_n}
\newcommand\numberthis{\addtocounter{equation}{1}\tag{\theequation}}

\begin{proof}
  We start by using the invariance of the RMC's steady state distribution probabilities, $\ss(a|s)$, when performing a reasoning step with transition probabilities from the BT-policy $\bt(a'|a,s)$. Hence, we can decompose $\g \E_{a\sim \pi_\theta^s(\cdot|s)}\left[Q^s(s, a)\right]$ in the sum of two distinct terms by applying the product rule:
  \begin{align*}
    \g &\E_{a\sim \ss(\cdot|s)}\left[Q^s(s, a)\right] = 
        \g \E_{a\sim \ss(\cdot|s)}\left[\E_{a'\sim \bt(\cdot|a, s)}\left[Q^s(s, a')\right]\right] %
    \\ &= \g \int_A \ss(a|s) \int_A \bt(a'| a, s) Q^s(s, a') da' da
    \\ &= \g \int_A \int_A \ss(a|s) \bt(a'| a, s) Q^s(s, a') da' da
    \\ &= \int_A \ss(a|s) \g \int_A \bt(a'| a, s) Q^s(s, a') da' da 
        + \int_A (\g\ss(a|s)) \int_A \bt(a'| a, s) Q^s(s, a') \annot{Leibniz integral rule}
    \\ &= \E_{a\sim \ss(\cdot| s)} \left[ \g \E_{a'\sim \bt(\cdot| a, s)}\left[Q^s(s, a')\right]\right]
        + \int_A (\g\ss(a|s)) \int_A \bt(a'| a, s) Q^s(s, a') da' da. \annot{(1) + (2)}
  \end{align*}
  From the definition of the n-step extensions being a \textit{local} approximation of the RMC with no dependence form $\theta$ (since $\g \qn(s,a)=\mathbf{0}$), we can rewrite term (2) as:
  \begin{align*}
    \int_A (\g\ss(a|s)) \int_A \bt(a'| a, s) Q^s(s, a') da' da. &= \g \int_A \ss(a|s) Q_1^s(s, a) da
    \\ &= \g \E_{a\sim \ss(\cdot|s)}\left[Q_1^s(s, a)\right].
  \end{align*}
  Hence, from the relationship between $\g \E_{a\sim \pi_\theta^s(\cdot|s)}\left[Q_0^s(s, a)\right]$ (i.e., $\g \E_{a\sim \pi_\theta^s(\cdot|s)}\left[Q^s(s, a)\right]$) with (1) and (2) we see that:
  \begin{equation*}
  \small
\g \E_{a\sim \ss(\cdot| s)}\left[Q^s_0 (s, a')\right] = \E_{a\sim \ss(\cdot| s)} \left[ \g \E_{a'\sim \bt(\cdot| a, s)}\left[Q^s_0(s, a')\right]\right]
        + \g \E_{a\sim \ss(\cdot| s)} \left[Q^s_1(s, a)\right]. \annot{(1) + (2)}
  \end{equation*}
  Since each n-step extension $Q^s_{n}$ has the same exact recursive relationship with the subsequent $Q^s_{n+1}$ n-step extension, more generally, we have that:
  \begin{equation} \label{eq:pr_rec_rel}
  \small
\g \E_{a\sim \ss(\cdot| s)}\left[Q^s_n (s, a')\right] = \E_{a\sim \ss(\cdot| s)} \left[\g \E_{a'\sim \bt(\cdot| a, s)}\left[Q^s_n(s, a')\right]\right]
        + \g \E_{a\sim \ss(\cdot| s)} \left[Q^s_{n+1}(s, a)\right].
  \end{equation}
  
  Thus, we can apply Equation~\ref{eq:pr_rec_rel} recursively to (2) and all resulting $\g \E_{a\sim \ss}\left[Q^s_n (s, a')\right]$ terms:
   \begin{align*}
  \g &\E_{a\sim \ss(\cdot| s)}\left[Q^s_0 (s, a')\right]= \E_{a\sim \ss(\cdot| s)} \left[\g \E_{a'\sim \bt(\cdot|a, s)}\left[Q^s_0(s, a')\right]\right]
        + \g \E_{a\sim \ss(\cdot| s)} \left[Q^s_1(s, a)\right]
  \\ &= \E_{a\sim \ss} \left[\g \E_{a'\sim \bt}\left[Q^s_0(s, a')\right]\right]
        + \E_{a\sim \ss} \left[ \g \E_{a'\sim \bt}\left[Q^s_1(s, a')\right]\right]
        + \g \E_{a\sim \ss} \left[Q^s_2(s, a)\right]
  \\ &= ...
  \\ &= \lim_{N\to \infty} \sum_{n=0}^{N} \E_{a\sim \ss} \left[\g \E_{a'\sim \bt}\left[Q^s_n(s, a')\right]\right]
        + \g \E_{a\sim \pi^s} \left[Q^s_N(s, a)\right]. \numberthis \label{eq:app:sum}
  \end{align*}
  Moreover, from the definitions of the n-step extensions $Q^s_n$ and the SS-policy $\ss$, we see that:
  \begin{align*}
     \lim_{N\to \infty} \g \E_{a\sim \ss} \left[Q^s_N(s, a)\right] &= \lim_{N\to \infty} \int_A (\g \ss(a|s)) Q^s_N(s, a) da \annot{Leibniz integral rule}
     \\ &= \int_A (\g \ss(a|s)) \lim_{N\to \infty} \int_A \pi^b_N(a'|a, s) Q^s(s,a') da'da
     \\ &= \int_A (\g \ss(a|s)) \int_A \ss(a'|s) Q^s(s,a') da'da
     \\ &= \g \int_A \ss(a|s) da \times \int_A \ss(a'|s) Q^s(s,a') da' \annot{Leibniz integral rule}
     \\ & = \mathbf{0} \numberthis \label{eq:app:toz}
  \end{align*}
  Therefore, we can use the identity from \ref{eq:app:toz} to simplify Equation~\ref{eq:app:sum}, and we are left with:
  \begin{align*} \label{eq:series_result}
      \g \E_{a\sim \ss(\cdot| s)}\left[\E_{a'\sim \bt(\cdot| a, s)}\left[Q^s (s, a')\right]\right] &=
            \lim_{N\to \infty} \sum_{n=0}^{n} \E_{a\sim \ss} \left[\g \E_{a'\sim \bt}\left[Q^s_n(s, a')\right]\right].
    \\ &=\E_{a\sim \ss} \left[\lim_{N\to \infty} \sum_{n=0}^{N} \g \E_{a'\sim \bt}\left[Q^s_n(s, a')\right]\right].
  \end{align*}
  
 \end{proof}

\textbf{Convergence and finite approximations.} We can use the results derived in Lemma~\ref{lemma:steady_state} (see \ref{app:subsec:app:lemma_p}) to establish some of the properties of the infinite series introduced in Theorem~\ref{thm:BT-POLGRAD}. To simplify notation, we will denote each term in the series from Equation~\ref{Eq.thm_eq} with $g_n(s, a)$, i.e.:

{\small
\begin{equation*} %
g_n(s, a) = \g \E_{a'\sim \pi^b_\theta(\cdot|a, s)} \left[Q^s_n(s, a')\right], 
\quad \text{such that }\g \E_{a\sim \pi_\theta^s}\left[Q^s(s, a)\right] = \E_{a\sim \pi_\theta^s}\left[\lim_{N\to \infty} \sum_{n=0}^{N} g_n(s,a)\right].
\end{equation*}
}%
We rewrite the value of each $g_n(s,a)$ term with the \textit{score function}, using the identity $\g \log(\bt(a'|a,s)) = \frac{\g \bt(a'|a,s)}{\bt(a'|a,s)}$:
\begin{align*} %
g_n(s, a) &= \g \E_{a'\sim \pi^b_\theta(\cdot|a, s)} \left[Q^s_n(s, a')\right]
\\ &= \int \g \pi^b_\theta(\cdot|a, s) Q^s_n(s, a') da'\annot{Leibniz integral rule}
\\ &= \int \pi^b_\theta(\cdot|a, s) \g \log(\bt(a'|a,s)) Q^s_n(s, a') da'
\\ &= \E_{a'\sim \bt(\cdot|a, s)} \left[\g \log(\bt(a'|a,s)) Q^s_n(s, a')\right].
\end{align*}
 Then, we explicitly subtract a baseline, $\E_{a''\sim \ss}[Q^s(s,a'')]$, from the n-step Q-function value $Q^s_n(s, a')$ multiplying the score. Note, that $\E_{a''\sim \ss}[Q^s(s,a'')]$ corresponds to the \textit{value function} of the SS-policy, which is a baseline \textit{independent} of $a'$. Therefore, $\E_{a'\sim \bt}\left[\g \log(\bt(a'|a,s))\E_{a''\sim \ss}[Q^s(s,a'')]\right] = 0$, showing that this subtraction leaves all the gradient terms unchanged \citep{suttonbarto}:
{\small
\begin{align*} %
g_n(s, a) &= \E_{a'\sim \bt(\cdot|a, s)} \left[\g \log(\bt(a'|a,s)) Q^s_n(s, a')\right]
    \\ &= \E_{a'\sim \bt(\cdot|a, s)} \left[\g \log(\bt(a'|a,s))\left(Q^s_n(s, a') - E_{a''\sim \ss(\cdot|s)}\left[Q^s\left(s,a''\right)\right]\right)\right]
    \\ &= \E_{a'\sim \bt(\cdot|a, s)} \left[ \g \log(\bt(a'|a,s)) \left(\int \pi^b_n(a''|a', s)Q^s(s,a'')da''- \int \ss(a''|s)Q^s(s,a'')da'' \right)\right]
    \\ &= \E_{a'\sim \bt(\cdot|a, s)} \left[\g \log(\bt(a'|a,s)) \int \left(\pi^b_n(a''|a', s) - \ss(a''|s)\right)Q^s(s,a'')da''\right] \numberthis \label{eq:app:drift_result}
\end{align*}
}%
From our boundedness assumptions, we define existing positive real values $Q^{+}$, $S^{+}$ such that:
\begin{gather*}
    Q^{+} \geq |Q^s(s,a)|, \quad S^{+} \geq | \g \log(\bt(a'|a,s))|,\quad \text{for all } a,a'\sim A, s\sim S.
\end{gather*}
Thus, by noting the relationship 
\begin{equation}
    \int \left|\pi^b_n(a''|a', s) - \ss(a''|s)\right| da'' \leq 2\times ||\pi^b_n(a''|a', s) - \ss(a''|s)||_{TV},
\end{equation}
we use the bound defined in Equation~\ref{eq:app:exp_conv_dtv} from Lemma~\ref{lemma:steady_state} to show that:
{\small
 \begin{align*} %
 g_n(s, a) &= \E_{a'\sim \bt(\cdot|a, s)} \left[\g \log(\bt(a'|a,s)) \int \left(\pi^b_n(a''|a', s) - \ss(a''|s)\right)Q^s(s,a'')da''\right]
    \\ &\leq \E_{a'\sim \bt(\cdot|a, s)} \left[\g \log(\bt(a'|a,s)) \int \left|\pi^b_n(a''|a', s) - \ss(a''|s)\right|Q^{+} da''\right]
    \\ &\leq \E_{a'\sim \bt(\cdot|a, s)} \left[ 2S^{+}Q^{+}||\pi^b_n(a''|a', s) - \ss(a''|s)||_{TV}\right]
    \\ &\leq 2S^{+}Q^{+}(1-\delta\times|A|)^n. \numberthis \label{eq:app:bound_result}
\end{align*}
 }%
 From Equation~\ref{eq:app:bound_result}, it is clear that each term in the series from Theorem~\ref{thm:BT-POLGRAD} will converge \textit{exponentially fast}. This shows we can estimate $\E_{a\sim \pi_\theta^s}\left[\lim_{N\to \infty} \sum_{n=0}^{N} g_n(s,a)\right]$ with any arbitrarily small error using a finite $N$. Moreover, we can see that each gradient term in Equation~\ref{eq:app:drift_result} is based on the expected deviation of the \textit{n-step} transition function from the steady-state distribution of the RMC. This property provides concrete intuition for the proposed adaptive truncation strategy, which considers the expected number of steps before reaching \textit{approximate convergence} (i.e., $\pi^b_N\approx \ss$). A similar gradient estimator for the steady-state distribution was also derived with more formal notation in prior works \citep{gradientEst_ss_pflug}. \citet{gradientEst_ss_pflug_unb_ext} further extended similar convergence results to broader classes of Markov chains and even certain unbounded performance functions, providing further motivation for our practical method.

\subsection{Policy gradient estimation for arbitrary regularized objectives}

We can extend Theorem~\ref{thm:BT-POLGRAD} to estimate the policy gradient with respect to a wider class of objectives that involve the expectation over \textit{regularized functions}:
\begin{equation}\label{eq:app:reg_obj_1}
    J(\theta) = \E_{s, a\sim \pi^s_\theta(\cdot|s)}\left[Q^r_\theta(s, a)\right],
\end{equation}
where the values of $Q^r_\theta(s,a)$ might depend on the parameters of the BT-policy, $\bt$. We assume $Q^r_\theta$ and its gradients respect the same regularity assumptions of continuity and boundedness stated in Theorem~\ref{thm:BT-POLGRAD}. The MaxEnt objective from Equation~\ref{eq:pg_bt_obj} is a particular instance of this setting with the value of $Q^r_\theta$  being dependent on the policy's entropy, i.e., $Q^r_\theta(s, a) = Q^s(s,a) - \alpha \log(\pi^s_\theta(a|s))$. Analogously to the unregularized case, we define a set of \textit{n-step extensions} for $Q^r_\theta$ for all $n=0,1,2,\dots$:
\begin{gather} \label{Eq:app:QREGnstepext}
Q^r_n(s, a) = \int_A \pi_n^b(a'|a,s) Q^r_\phi(s,a') da', \quad 
\textit{with} \quad\g Q^r_n(s, a) = \mathbf{0}.
\end{gather}
For any state $s$, we can then apply the product rule to rewrite the gradient of Equation~\ref{eq:app:reg_obj_1} as a sum of two expectations:
\begin{align*}\label{eq:app:pred_reg_split}
    \g J(\theta) &= \g \E_{a\sim \pi^s_\theta(\cdot|s)}\left[Q^r_\theta(s, a)\right]
    \\ &=  \int_A (\g \ss(a|s)) Q^r_\theta(s, a') da +  \int_A \ss(a|s) \g Q^r_\theta(s, a') da. \annot{Leibniz integral rule}
    \\ &=  \g \E_{a\sim \pi^s_\theta(\cdot|s)}\left[Q^r_0(s, a)\right] + \E_{a\sim \pi^s_\theta(\cdot|s)}\left[\g Q^r_\theta(s, a)\right]. \annot{(1) + (2)}
\end{align*}
Here, we used the definition of each $Q^r_n(s, a)$ term being a \textit{local approximation} and having no dependency on the BT-policy to rewrite term (1) as an expectation. Hence, we can directly apply the unregularized version of Theorem~\ref{thm:BT-POLGRAD} to term (1), yielding:
\begin{align*}
    \g J(\theta) =  \E_{a\sim \ss}\left[\lim_{N\to \infty} \sum_{n=0}^{N} \g \E_{a'\sim \bt}\left[Q^r_n(s, a')\right]\right] + \E_{a\sim \ss}\left[\g Q^r_\theta(s, a)\right]. \annot{(1) + (2)}
\end{align*}
Since $\E_{a\sim \ss}\left[\g Q^r_\theta(s, a)\right]=\E_{a\sim \ss}\left[\E_{a'\sim \bt(\cdot|a,s)}\left[\g Q^r_\theta(s, a')\right]\right]$, we can merge back the two expectations, obtaining:
\begin{align*}\label{eq:app:thm_to_unreg}
    \g J(\theta) &= \E_{a\sim \ss}\left[\g Q^r_\theta(s, a') + \lim_{N\to \infty} \sum_{n=0}^{N} \g \E_{a'\sim \bt}\left[Q^r_n(s, a')\right]\right]
    \\ &=  \E_{a\sim \ss}\left[\g \E_{a'\sim \bt}\left[Q^r_\theta(s, a')\right] + \lim_{N\to \infty} \sum_{n=1}^{N} \g \E_{a'\sim \bt}\left[Q^r_n(s, a')\right]\right]. \numberthis
\end{align*}
The resulting form of the policy gradient generalizes the original Equation~\ref{Eq.thm_eq} from Theorem~\ref{thm:BT-POLGRAD}, where now the \textit{first term} in the infinite series explicitly considers the full gradients from the regularized objective. Hence, applying truncation and the reparameterization trick (as in Equation~\ref{Eq.whole_reparam}) now yields:%
\small
\begin{gather*}
        \small
    \g J(\theta) \approx \E_{\epsilon_0,\dots,\epsilon_N}\left[\g Q^r_{\theta}(s,a_1) + \sum^N_{n=0} \nabla_{a_n} Q^r_{\theta}(s,a_n) \left(\prod^{n-1}_{i=0} \nabla_{a_i} f^b(a_{i}, s, \epsilon_{i+1})\right)\g f_\theta^b(a, s, \epsilon_0)\right], \\
    \text{where } a\sim \ss(\cdot|s), \quad a_0=f^b(a, s, \epsilon_{0}), \quad \text{and } a_i=f^b(a_{i-1}, s, \epsilon_{i})\quad  \text{for all } i=1,\dots,n. \numberthis \label{eq:app:reg_obj}
\end{gather*}
\normalsize

\textbf{Practical considerations for MaxEnt RL.} The reparameterized extension in Equation~\ref{eq:app:reg_obj} is directly analogous to the extension of DDPG policy gradients \citep{ddpg} in SAC \citep{sac, sac-alg} when considering the MaxEnt regularized objective:
\begin{equation}
    Q^r_\theta(s, a) = Q^{s}(s,a) - \alpha \log\pi^s_{\theta}(a|s), \quad \g Q^r_\theta(s, a) = -\g \alpha \log\pi^s_{\theta}(a|s).
\end{equation}
To estimate the entropy of the SS-policy, we use the approximation $\ss(a|s)=\E_{\ss(a'|s)}\left[\bt(a|s)\right]\approx \frac{1}{N+1}\sum^N_{n=0}\bt(a|a_n,s)$, which results in a \textit{nested Monte-carlo estimator} \citep{nested-MC-estimator} for evaluating this component of the policy gradient with reparameterization. We find that using $N=\ceil{\hat{N}}$ (Section~\ref{subsec:ss_conv_gelrub}) ensures enough samples for practical effectiveness, without necessitating additional debiasing tricks (e.g., \citep{MC-debiasing-trick}).

\subsection{Unlimited expressivity}

The unlimited representation power of Serial Markov Chain Reasoning comes from the fact that the distribution of agent behavior given by the SS-policy, $\ss$, is a mixture model with potentially infinitely many components. Hence, even a simple Gaussian parameterization of the BT-policy $\bt$ makes such distribution an arbitrary mixture of Gaussian distributions, which enables agent behavior to approximate any action distribution to arbitrary precision \citep{FiniteMixStatAna}. This is due to Gaussian mixtures being universal approximators of densities \citep{lecun_deeplearningbook}. Recent works also show that the same property extends to arbitrary mixture models, with implications for future non-Gaussian extensions of our framework \citep{nestoridis_universalseries}.

\newpage

\section{Convergence detection}

\label{app:gel_rub_impl}

\begin{figure}[H]
    \centering
    \includegraphics[width=0.7\linewidth]{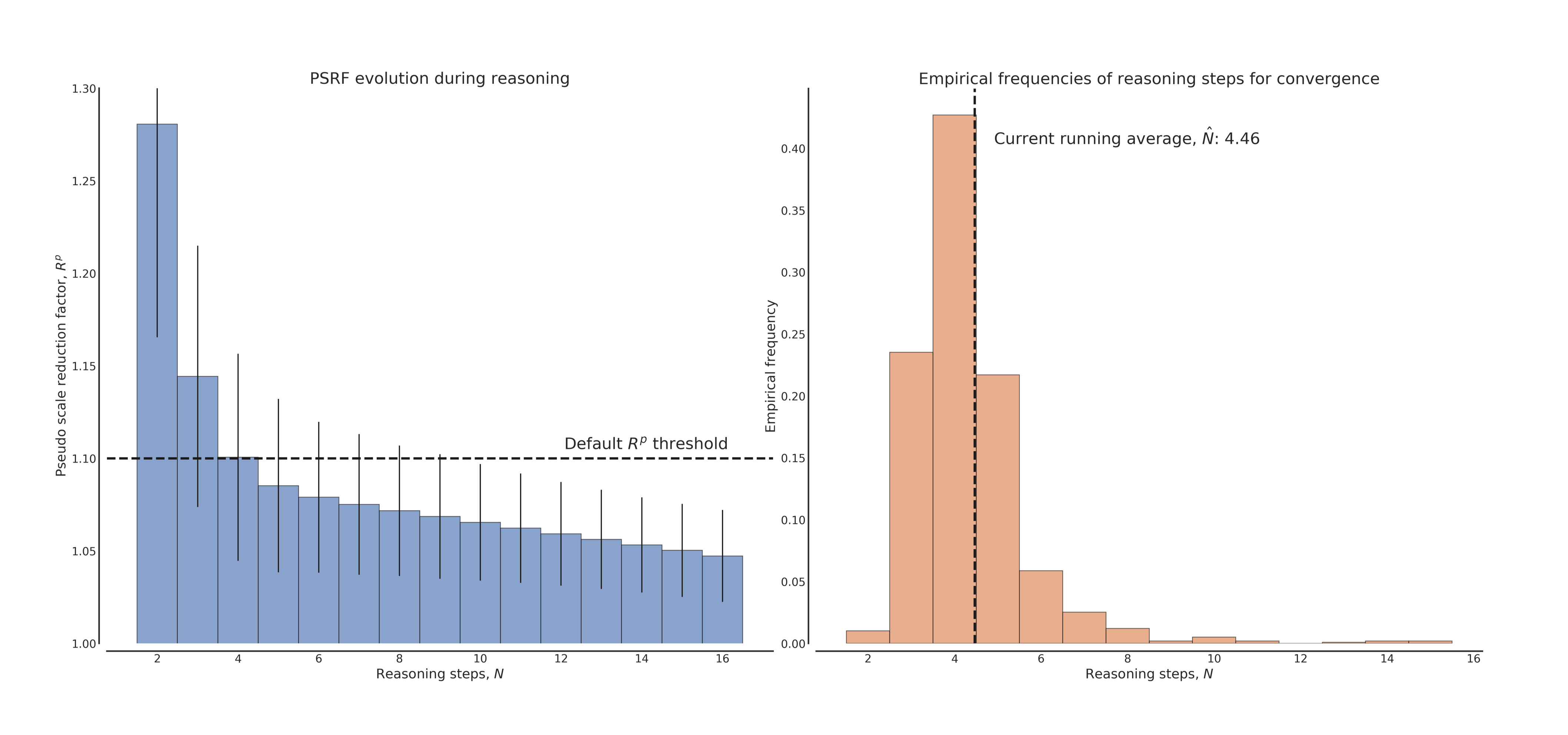}
    \caption{Pseudo scale reduction factor statistics collected after training an agent in the Ant-v2 environment from OpenAI Gym Mujoco with SSPG. (\textbf{Left}) Evolution of the average PSRF throughout an episode by performing up to 16 reasoning steps for each action selection decision. (\textbf{Right})) Empirical frequencies of the reasoning steps performed before approximate convergence to the steady-state distribution of the RMC, as determined by $R^p<1.1$.}
    \label{fig:app:BCOMB_psrf}
\end{figure}%
As described in Section~\ref{subsec:ss_conv_gelrub}, we detect convergence in the reasoning process using the multivariate version of the seminal Gelman-Rubin (GR) convergence diagnostic \citep{gelmanrubin, gelmanrubin-gen-mult}. Given a set of $M$ parallel reasoning chains of length $N$, $a_{1:N}$, this diagnostic computes the \textit{pseudo scale reduction factor} (PSRF), a score that should expectedly converge to 1 as $N$ increases and the distribution of the different chains approaches the steady-state distribution. We provide an example in Figure~\ref{fig:app:BCOMB_psrf} Left where we display how the average PSRF evolves after performing 2-16 reasoning steps after training an agent for the Ant-v2 Mujoco task. As described in the main text, the PSRF ($R^p$) is based on the largest eigenvalue from the matrix product of the sample covariance \textit{within} ($W$) each of the parallel chains ($\left[a^i_1,\dots,a^i_M\right]$, for $i=1,2,\dots, M$) and an unbiased estimate of the true covariance of the steady-state distribution, utilizing the sample covariance \textit{between} (B) the different chains:
\begin{equation*}
    R^p = \sqrt{\frac{N-1}{N} + \lambda_{max}(W^{-1}B)}.
\end{equation*}
This score can be interpreted as the largest variability in any direction between these two estimates, which can be much higher than the ratio of individual variances computed by the univariate GR scores \citep{gelmanrubin-revisited}. Intuitively, as the variance within each chain approaches the variance between independent chains, we can take this as a strong indication that we sufficiently explored the Markov chain's state-space within each chain. For our problem settings, we found the PSRF to work well with the default convergence threshold $R^p < 1.1$. After detecting convergence, we use the number of performed reasoning steps ($N$) to update a running mean $\hat{N} = \rho\hat{N} + (1-\rho)N$, as shown in Algorithm~\ref{alg:act_sel}. To perform this update, we found a `untuned' choice of $\rho=0.99$ to also work appropriately. We observed that for any individual task and training iteration, the majority of the agent's action selection decisions requiring a number of steps close to $\hat{N}$, as shown by the example in Figure~\ref{fig:app:BCOMB_psrf} Right. Hence, rather than computing the PSRF at every step we found it more efficient to compute the first score only after simulating the RMC for $N=\floor{\hat{N}}$ reasoning steps. In particular, starting from $\mathbf{a_{1:\floor{\hat{N}}}}$, we consider two cases: If $R^p \geq 1.1$, we simulate and compute the PSRF for all further steps until convergence. Otherwise, if $R^p < 1.1$, we backtrack to find the minimum $N$, such that the subsequence $\mathbf{a_{1:N}}$ still satisfies our threshold. With this strategy, we effectively limit the number of evaluations of the PSRF and bolster overall efficiency.

\textbf{Considerations.} While the assumptions underlying the properties of the GR convergence diagnostic \citep{gelmanrubin, gelmanrubin-gen-mult} are hard to ensure in practical scenarios, we believe its widespread adoption and empirical effectiveness are strong motivators for its use. Recent extensions \citep{gelmanrubin-revisited} incorporated modern techniques for variance/covariance estimation together with less conservative termination criteria. Such methods could provide further efficiency gains to our framework, but might also introduce alternative forms of bias while acting and learning. We leave such exploration and analysis to future work.

\newpage

\section{Implementations details}

\label{app:sspg_impl}

In this Section, we provide descriptions of the SSPG implementations and experimental setups, together with comprehensive hyper-parameters lists. Since serial Markov chain reasoning is a novel framework, where action-selection is an adaptive iterative process, it adds additional implementation complexity compared to traditional reinforcement learning. Hence, we share our implementation to ensure reproducibility and facilitate future extensions/comparisons. By default, the only main requirement of implementing the \textit{serial Markov chain reasoning} framework is to substitute traditional policies $\pi(\cdot|s)$ with BT-policies $\pi^b(\cdot|a, s)$, which take an additional input from the agent's action space to perform \textit{reasoning}. In the considered implementations, we also store a \textit{short-term action memory buffer}, $\hat{A}$, for acting in the environment, as described in Section~\ref{subsec:action_sel_temp_cons}. Overall, we mostly re-use hyper-parameters and practices from existing policy gradient implementations without major SSPG-specific tuning. In total, our algorithm only introduces 4 new main hyper-parameters: 1. The number of initial action beliefs, $M$. 2. The pseudo scale reduction factor (PSRF) threshold to detect approximate convergence. 3. The size of the short-term memory action buffer, $\hat{A}$. 4. The coefficient for updating the running mean of performed reasoning steps, $\rho$ (i.e., to update the value of $\hat{N}$ for learning). The values for these new hyper-parameters were mostly chosen based on default values from similar practices and preliminary experimentation. Overall, SSPG's performance appeared to be quite robust to a range of reasonable choices and we kept them \textit{fixed} for all tested domains. We provide ablation studies and examine performance with different settings in Appendix~\ref{app:abl}. These results further confirm the claimed robustness.

\subsection{OpenAI Gym Mujoco}

\begin{table}[H]
\caption{SSPG hyper-parameters on OpenAI Gym Mujoco tasks} 
\label{tab:app:mj_hyper}
\begin{center}
\begin{tabular}{ll} 
\toprule
\multicolumn{2}{c}{General MaxEnt RL hyper-parameters (following common practices \citep{mbpo, redq}}                                                                                                                              \\ 
\midrule
Replay data buffer size                                                      & 1000000                                                                                                                                                    \\
Batch size                                                                   & 256                                                                                                                                                        \\
Minimum data before training                                                 & 5000                                                                                                                                                       \\
Random exploration steps                                                     & 5000                                                                                                                                                       \\
Optimizer                                                                    & \textit{Adam} \cite{adam}                                                                                                                                         \\
Policy/critic learning rate                                                  & 0.0003                                                                                                                                                     \\
Policy/critic $\beta_1$                                                      & 0.9                                                                                                                                                        \\
Critic UTD ratio                                                             & 10 (half the other baselines)                                                                                                                              \\
Policy UTD ratio                                                             & 1                                                                                                                                                          \\
Discount $\gamma$                                                            & 0.99                                                                                                                                                       \\
Target critic polyak coefficient                                                  & 0.995                                                                                                                                                      \\
Hidden dimensionality                                                        & 256                                                                                                                                                        \\
Fully-connected hidden layers                                                       & 3                                                                                                                                                        \\
Nonlinearity                                                                 & ReLU                                                                                                                                                       \\
Initial entropy coefficient $\alpha$                                         & 1                                                                                                                                                          \\
Entropy coefficient learning rate                                            & 0.0001                                                                                                                                                     \\
Entropy coefficient $\beta_1$                                                & 0.5                                                                                                                                                        \\
Policy target entropy                                                        & \begin{tabular}[c]{@{}l@{}}\textit{Hopper} : -1, \textit{HalfCheetah}: -3, \\ \textit{Walker2d}: -3, \textit{Ant}: -4, \textit{Humanoid}: -2\end{tabular}  \\
Critic ensemble size                                                         & 10                                                                                                                                                         \\
Critic penalization (uncertainty regularizer \citep{learningpessimism}) & 0.75                                                                                                                                                       \\ 
\midrule
\multicolumn{2}{c}{SSPG-specific hyper-parameters}                                                                                                                                                                                        \\ 
\midrule
Number initial action beliefs, $M$                                          & 64                                                                                                                                                         \\
PSRF threshold                                                               & 1.1                                                                                                                                                        \\
Short-term action memory size                                                & 64                                                                                                                                                         \\
Reasoning steps moving average coefficient, $\rho$          & 0.99                                                                                                                                                       \\
\bottomrule
\end{tabular}
\end{center}
\end{table}

Our Mujoco SSPG implementation follows common practices employed in the considered baselines and recent literature. Following REDQ \citep{redq} and MBPO \citep{mbpo}, we employ a critic ensemble of 10 models and use the suggested task-specific target entropy values for automatic tuning of the MaxEnt coefficient, $\alpha$ \citep{sac-alg}. We parameterize each critic with a \textit{modern} architecture \citep{deeper-deep-RL} employing 3 fully connected hidden layers as in \citep{learningpessimism}. Inline with observations for IAPO \citep{iterativeAmortizedPolOptim}, we find that performing iterative computations with the policy model can easily lead to a value overestimation phenomenon. Hence, to compute the Q-function's targets, we use the uncertainty regularizer \citep{learningpessimism} with a slightly-increased fixed coefficient of 0.75 (the default value of 0.5 is equivalent to the penalization from clipped double Q-learning \citep{td3}). See Table~\ref{tab:app:mj_hyper} for a full list of hyper-parameters.

\subsection{DeepMind Control}

\begin{table}[H]
\caption{SSPG hyper-parameters on DeepMind Control tasks from pixels} 
\label{tab:app:dmc_hyper}
\begin{center}
\begin{tabular}{@{}ll@{}}
\toprule
\multicolumn{2}{c}{Pixel-based RL hyper-parameters (same as DrQv2 \citep{drqv2})}                   \\ \midrule
Replay data buffer size                            & 1000000 (quadruped\_run: 100000)           \\
Batch size                                         & 256 (walker\_run: 512)                     \\
Minimum data before training                       & 4000                                      \\
Random exploration steps                           & 2000                                      \\
Optimizer                                          & \textit{Adam} \citep{adam}                \\
Policy/critic learning rate                        & 0.0001                                    \\
Policy/critic $\beta_1$                            & 0.9                                       \\
Critic UTD ratio                                   & 0.5                                       \\
Policy UTD ratio                                   & 0.5                                       \\
Discount $\gamma$                                  & 0.99                                      \\
Target critic polyak coefficient                   & 0.99                                      \\
N-step                                             & 3 (walker\_run: 1)                         \\
Convolutional kernel size                          & 3$\times$3                               \\
Convolutional filters                              & 32                                        \\
Convolutional layers                               & 4                                         \\
Convolutional strides                              & 2, 1, 1, 1                                \\
Feature dimensionality                             & 50                                        \\
Hidden dimensionality                              & 256                                       \\
Fully-connected hidden layers                      & 3                                         \\
Nonlinearity                                       & ReLU                                      \\ \midrule
\multicolumn{2}{c}{General MaxEnt RL hyper-parameters}                                         \\ \midrule
Initial entropy coefficient $\alpha$               & 1                                         \\
Entropy coefficient learning rate                  & 0.0001                                    \\
Entropy coefficient $\beta_1$                      & 0.5                                       \\
Policy target entropy                              & 0.5$\to-$1$\times$ dim($A$) in 500K steps \\ \midrule
\multicolumn{2}{c}{SSPG-specific hyper-parameters}                                             \\ \midrule
Number initial action beliefs, $M$                 & 64                                        \\
PSRF threshold                                     & 1.1                                       \\
Short-term action memory size                      & 64                                        \\
Reasoning steps moving average coefficient, $\rho$ & 0.99                                      \\ \bottomrule
\end{tabular}
\end{center}
\end{table}

Our DeepMind Control pixel-based SSPG implementation mostly follows the exact hyper-parameters and model specifications from DrQv2 \citep{drqv2}, but re-introduces MaxEnt RL in the policy gradient objectives. The policy target entropy is annealed from $0.5\times dim($A$)$ to $-1\times dim($A$)$ in the first 500K steps, which was done to mimic the exploration standard deviation annealing (from 1 to 0.2 in the first 500K steps) again from the DrQv2 implementation. Both the other MaxEnt RL parameters and SSPG-specific parameters are kept identical to our experiments on Mujoco tasks. See Table~\ref{tab:app:dmc_hyper} for a full list of hyper-parameters.

\subsection{Positional bandits}

To provide visualizations and intuition for the behavior of RL agents adopting the serial Markov chain reasoning framework, we design simple few-dimensional \textit{toy tasks}, which we call \textit{positional bandits}. We used a 1-dimensional positional bandit for the example in Figure~\ref{fig:bt_ss_pol} and 2-dimensional positional bandits for our expressiveness experiments in Section~\ref{subsec:4prop_exps}. Positional bandits are defined by an arbitrary list of goal coordinates within a bounded state-space centered around 0. The action space represents position coordinates, determining which part of the state-space the agent will immediately visit. The reward of each action is proportional to the distance it brings the agent to the closest goal coordinate. Thus, in each \textit{single-step episode}, the agent receives \textit{information} about its proximity to only one of the goals. In these environments, we train all agents with the MaxEnt objective, making optimal behavior correspond to visiting all-goals with similar frequencies. These minimal problems aim to isolate and assess the ability of different agents of modeling arbitrary multi-modal distributions of behavior in the \textit{non i.i.d.} RL problem setting which also involves a non-trivial exploration challenge.

We evaluate light-weight versions of SSPG, SAC, and a flow-based version of SAC. We largely follow the specifications from the Mujoco experiments, with a shared UTD of 1, a single critic, 50 random exploration steps, a fixed $\alpha$, and with all models (except the flow-based policy) using 2-layer fully-connected architectures with 32 hidden dimensions. We implemented the flow-based version of SAC, based on inverse autoregressive flows \citep{invAutoFlows}, as also considered in related RL works \citep{PG-norm-flows, iterativeAmortizedPolOptim}. In particular, we apply to the policy model's outputs two additional flow transformations, each parameterized by a 2-layer fully connected network. We found that this class of models benefits from wider policy networks, thus, we increased back the hidden dimensionality to 256 (as in our main Mujoco experiments). We found 1000 steps of experience collection/network updates was sufficient for behavior to convergence in SSPG and standard SAC. Instead, the flow-based version of SAC required much additional experience/training before attaining behavior resembling any actual multi-modality, hence, we considered its results after 1M steps.

\subsection{Other considerations}

In this work, we implemented SSPG with minimal extensions to existing policy gradient models and parameters. The purpose of this choice was to remove confounding factors related to hyper-parameter and model tuning, and evaluate our framework mainly for its conceptual properties over traditional RL agents. For the same reason, we concentrated our empirical analysis on established evaluation benchmarks and toy problems, rather than exploring new challenging applications. This leaves much potential for future work to investigate new model architectures, training practices, and push the limit of \textit{serial Markov chain reasoning}.

Early versions of our framework considered using some recurrent memory components as input to the BT-policy. However, to preserve the Markov property, this modification would require considering an extended state space for the RMC, consisting of both action-beliefs and memory hidden states. Preliminary results showed that such extended state space makes convergence require an increased amount of reasoning steps and optimization of the BT-policy more unstable.

\newpage

\section{Extended evaluation results}

\label{app:ext_res}

In this Section, we provide granular per-task results and further aggregate evaluations with the \textit{Rliable} evaluation protocol~\citep{precipice-rliable} after collecting different numbers of steps. We also compare the performance of SSPG with a new baseline integrating prior state-of-the-art algorithms with the normalizing flow model introduced in the first part of Section~\ref{subsec:4prop_exps}. Additionally, we report the mean running times for the considered algorithms and the main baselines we run to obtain performance results. Together with details about our hardware setup, these should give a solid intuition of the relative computational training cost of each algorithm. Furthermore, we also provide results for the performance of SSPG when `clipping' the maximum number of reasoning steps allowed for each action-selection, and the relative evaluation times to analyze potential deployment-time efficiency and trade-offs.

\subsection{OpenAI Gym Mujoco}

\begin{table}[H]
\tabcolsep=0.05cm
\caption{Per-task results for the considered OpenAI Gym Mujoco tasks. Each displayed return is averaged over 5 random seeds from 100 test rollouts from the preceding 10K training steps.} \label{tab:mj_perf}
\centering
\adjustbox{max width=0.98\linewidth}{

\begin{tabular}{lcccccccccc} 
\toprule
                    & \multicolumn{5}{c}{100K steps}                                          & \multicolumn{5}{c}{200K steps}                                            \\ 
\cmidrule(lr){2-6}\cmidrule(lr){7-11}
Tasks               & \textbf{SSPG}     & REDQ      & IAPO     & MBPO              & SAC-20   & \textbf{SSPG}     & REDQ      & IAPO     & MBPO               & SAC-20    \\ 
\cmidrule(lr){1-1}\cmidrule(lr){2-6}\cmidrule(r){7-11}
Invertedpendulum-v2 & 1000±0            & 1000±0    & 1000±0   & 963±37            & 1000±0   & 1000±0            & 1000±0    & 1000±0   & 963±37             & 1000±0    \\
Hopper-v2           & \textbf{3314±68}  & 2994±510  & 425±229  & 3271±192          & 2718±908 & \textbf{3487±87}  & 3060±617  & 426±149  & 3303±203           & 3356±26   \\
Walker2d-v2         & \textbf{4428±230} & 1989±1003 & 476±107  & 3393±528          & 2043±757 & \textbf{4793±186} & 2969±861  & 570±74   & 4034±485           & 3039±903  \\
Halfcheetah-v2      & 8897±496          & 5613±436  & 4122±566 & \textbf{9533±332} & 5831±723 & 10309±653         & 6633±568  & 5303±597 & \textbf{10670±750} & 7187±839  \\
Ant-v2              & \textbf{5163±275} & 3132±1243 & 5±19     & 1596±446          & 496±105  & \textbf{5513±238} & 3792±1064 & 118±83   & 4309±632           & 1801±776  \\
Humanoid-v2         & \textbf{4992±140} & 1402±657  & 441±90   & 559±62            & 495±96   & \textbf{5148±51}  & 4721±648  & 390±160  & 3316±774           & 405±200   \\
\bottomrule
\end{tabular}

}
\end{table}

\textbf{Per-task results.} In Table~\ref{tab:mj_perf} we provide the per-task results collected at 100K and 200K steps for the considered OpenAI Gym Mujoco tasks. For both experience thresholds, SSPG obtains the best average performance in 5/6 tasks, and still lags very close the model-based MBPO~\citep{mbpo} algorithm for the remaining task (HalfCheetah-v2). SSPG converges much earlier than other algorithms, even while performing many less optimization steps (REDQ, REDQ-FLOW, MBPO, and SAC-20 all employ a UTD of 20, while we train SSPG with a UTD of 10, see Section~\ref{sec:4exps}).

\begin{figure}[H]
    \centering
    \includegraphics[width=0.7\linewidth]{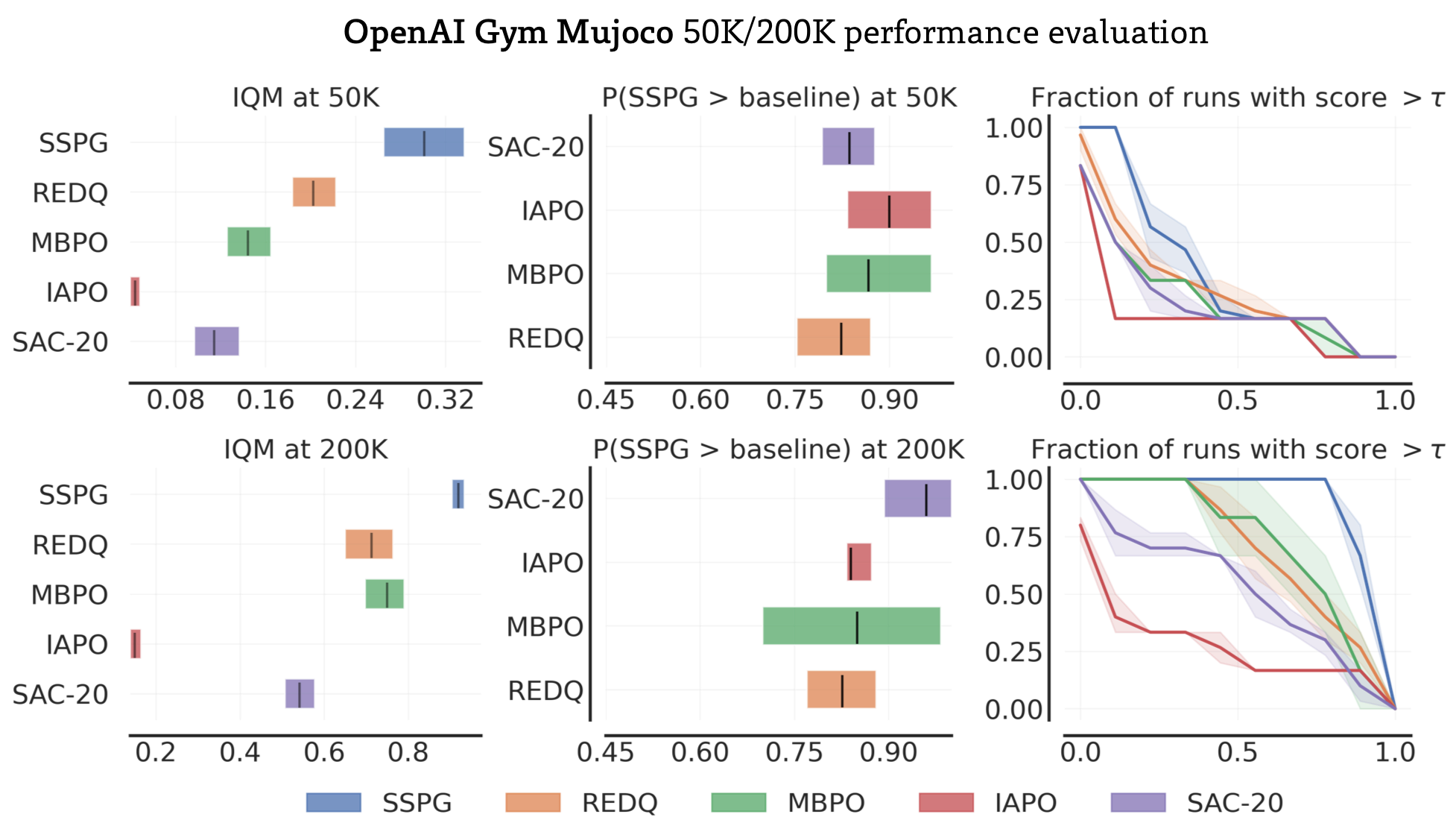}
    \caption{OpenAI Gym Mujoco aggregate performance evaluation with \textit{Rliable} \citep{precipice-rliable} after 50K (\textbf{Top}) and 200K (\textbf{Bottom}) training steps. These results complement the analogous results after 100K training steps together with the performance curves provided in Section~\ref{sec:4exps}.}
    \label{fig:app:ext_rliable_mj}
\end{figure}%

\textbf{Extended aggregate results.} In Figure~\ref{fig:app:ext_rliable_mj}, we provide additional aggregate metrics collected at 50K and 200K steps, using the same statistical tools described in Section~\ref{sec:4exps}. We see most of the considerations for the evaluation at 100K steps, equally hold at these different experience regimes. Most notably, SSPG aggregately outperforms all prior algorithms with \textit{statistically meaningful} gains, for the Neyman-Pearson statistical testing criterion \citep{stat-significance-neyman-pearson} and also exhibits stochastic dominance \citep{stoch-dom} after 200K training steps.

\subsection{DeepMind Control}

\begin{table}[H]
\tabcolsep=0.05cm
\caption{Per-task results for the considered DeepMind Control tasks. Each displayed return is averaged over 5 random seeds from 100 test rollouts from the preceding 150K training steps.} \label{tab:dmc_perf}
\centering
\adjustbox{max width=0.98\linewidth}{

\begin{tabular}{lcccccccccc} 
\toprule
                          & \multicolumn{5}{c}{1.5M steps}                                   & \multicolumn{5}{c}{3M steps}                                     \\ 
\cmidrule(lr){2-6}\cmidrule(lr){7-11}
Tasks                     & \textbf{SSPG}    & DrQv2           & DrQ     & CURL    & SAC     & \textbf{SSPG}   & DrQv2           & DrQ     & CURL    & SAC      \\ 
\cmidrule(lr){1-1}\cmidrule(lr){2-6}\cmidrule(lr){7-11}
acrobot\_swingup          & 218±49           & \textbf{272±40} & 24±18   & 6±3     & 8±5     & 371±41          & \textbf{422±48} & 43±37   & 6±4     & 10±6     \\
cartpole\_swingup\_sparse & \textbf{797±43}  & 478±391         & 321±393 & 497±330 & 135±268 & \textbf{837±15} & 503±411         & 319±391 & 530±353 & 174±290  \\
cheetah\_run              & 755±47           & \textbf{781±32} & 777±63  & 520±106 & 8±6     & \textbf{888±10} & 873±55          & 832±31  & 589±93  & 7±7      \\
finger\_turn\_easy        & \textbf{794±127} & 757±156         & 184±63  & 305±107 & 170±60  & \textbf{974±6}  & 932±43          & 218±106 & 310±139 & 211±98   \\
finger\_turn\_hard        & \textbf{637±138} & 506±229         & 89±48   & 224±134 & 79±53   & \textbf{945±42} & 913±60          & 100±65  & 172±75  & 100±55   \\
hopper\_hop               & \textbf{246±28}  & 200±102         & 272±88  & 185±129 & 0±0     & \textbf{344±28} & 239±123         & 290±86  & 223±133 & 0±0      \\
quadruped\_run            & \textbf{570±22}  & 402±213         & 135±77  & 182±97  & 61±47   & \textbf{760±64} & 494±288         & 115±36  & 170±86  & 56±32    \\
quadruped\_walk           & \textbf{855±23}  & 591±271         & 147±131 & 126±41  & 68±52   & 888±22          & \textbf{905±44} & 124±40  & 154±31  & 52±28    \\
reach\_duplo              & \textbf{221±7}   & 219±7           & 9±15    & 10±11   & 0±1     & 218±9           & \textbf{228±1}  & 9±6     & 7±7     & 2±2      \\
reacher\_easy             & \textbf{978±4}   & 973±3           & 587±198 & 713±87  & 64±59   & \textbf{982±3}  & 954±22          & 600±163 & 645±156 & 100±61   \\
reacher\_hard             & \textbf{913±77}  & 802±113         & 343±242 & 482±179 & 7±15    & \textbf{974±6}  & 944±25          & 426±288 & 651±324 & 15±19    \\
walker\_run               & \textbf{634±16}  & 568±273         & 477±153 & 378±235 & 26±4    & \textbf{738±7}  & 616±297         & 549±139 & 449±223 & 26±4     \\ 
\cmidrule(lr){1-1}\cmidrule(lr){2-6}\cmidrule(lr){7-11}
Average score             & \textbf{634.79}  & 545.72          & 280.34  & 302.20  & 52.28   & \textbf{743.32} & 668.60          & 301.97  & 325.41  & 62.77    \\ 
\cmidrule(lr){1-1}\cmidrule(lr){2-6}\cmidrule(lr){7-11}
Median score              & \textbf{695.85}  & 537.37          & 228.13  & 264.37  & 43.49   & \textbf{862.73} & 744.66          & 253.97  & 266.34  & 39.08    \\
\bottomrule
\end{tabular}

}
\end{table}

\textbf{Per-task results.} In Table~\ref{tab:dmc_perf} we provide the per-task results collected at 1.5M and 3M steps for the considered DeepMind Control tasks from pixel observations from \citep{drqv2}. SSPG obtains the best average performance in 10/12 and 9/12 tasks, respectively. We can observe most notable gains in some of the tasks that pose harder exploration challenges (e.g., cartpole\_swingup\_sparse) and especially in lower-data regimes (e.g., finger\_turn\_hard). We believe this could be an indication that \textit{serial Markov chain reasoning} complements particularly well the MaxEnt reinforcement learning framework and is able to overcome some of its observed limitations for tackling sparse reward, pixel-based tasks \citep{drqv2}.

\begin{figure}[H]
    \centering
    \includegraphics[width=0.7\linewidth]{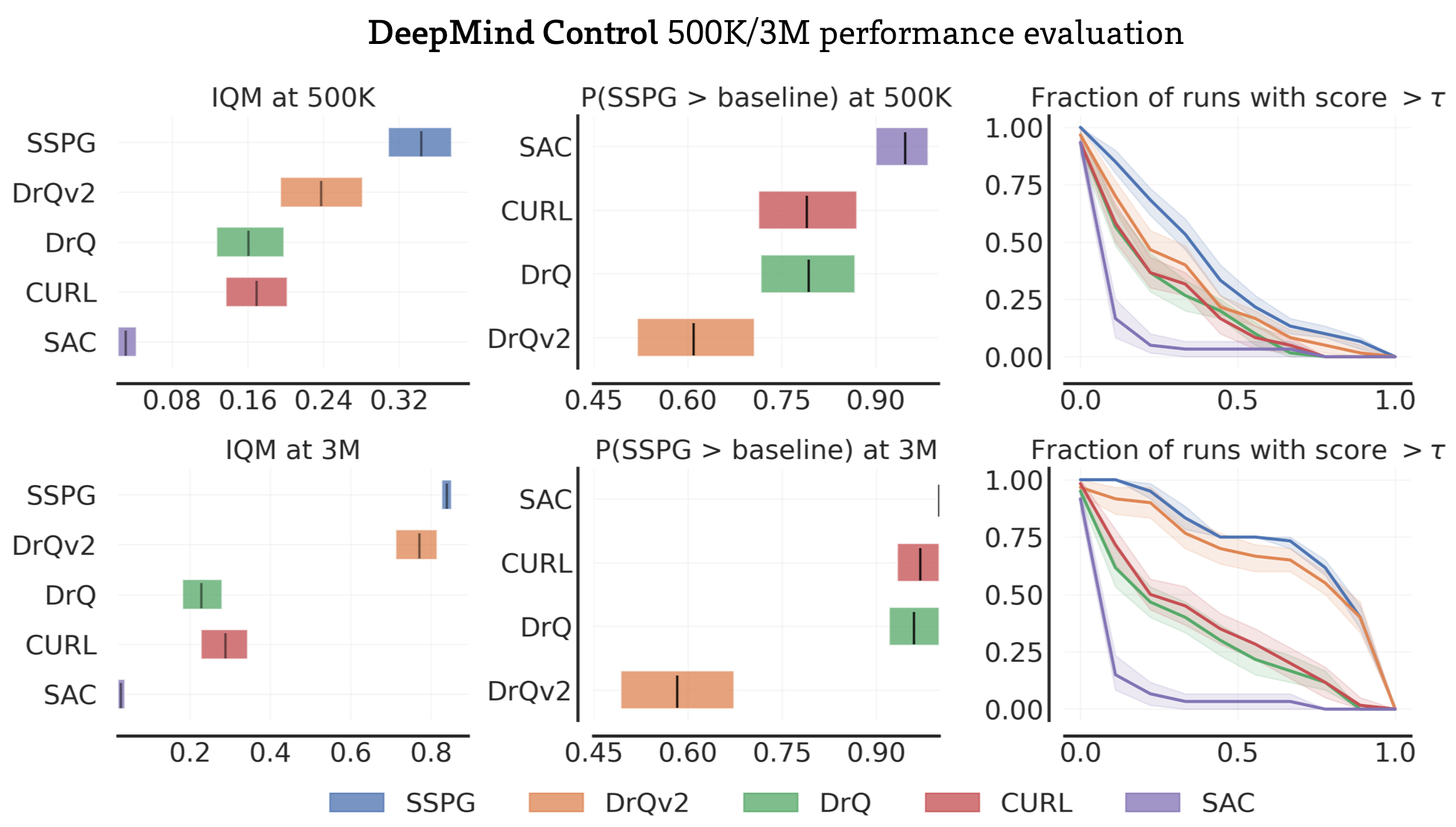}
    \caption{DeepMind Control aggregate performance evaluation with \textit{Rliable} \citep{precipice-rliable} after 500K (\textbf{Top}) and 3M (\textbf{Bottom}) training steps. These results complement the analogous results after 1.5M training steps together with the performance curves provided in Section~\ref{sec:4exps}.}
    \label{fig:app:ext_rliable_dmc}
\end{figure}%

\textbf{Extended aggregate results.} In Figure~\ref{fig:app:ext_rliable_dmc}, we provide additional aggregate metrics collected at 500K and 3M steps, using the same statistical tools described in Section~\ref{sec:4exps}. Again, these results are consistent with the evaluation at 1.5M steps. In particular, we highlight ubiquitous statistical significance and stochastic dominance \citep{stoch-dom} after training from 500K environment steps.

\subsection{Comparison with normalizing flows} %

We provide results comparing SSPG with additional new baselines, extending state-of-the-art algorithms based on the traditional RL framework with the normalizing flow policy model considered in Section~\ref{subsec:4prop_exps}. To counteract the exploration collapse phenomenon observed in the positional bandits experiments, we found it beneficial to add a small amount of fixed Gaussian noise on top of the already-stochastic policy when collecting training data ($\sigma=0.05$). While powerful flow models can also provide unlimited expressiveness, they are still based on the traditional RL framework and represent decision-making as the output of a \textit{fixed} process. Thus, these experiments can provide insights regarding the contribution of adaptivity to our SSPG framework. We compare the relative improvements from SSPG and flows with respect to the original performance of the considered state-of-the-art baselines in either the OpenAI Gym Mujoco and DeepMind Control benchmarks.

\begin{table}[H]
\tabcolsep=0.05cm
\caption{Per-task results comparing improvements from SSPG and normalizing flows over REDQ for the considered OpenAI Gym Mujoco tasks. The results were collected as described in Table~\ref{tab:mj_perf}. }
\label{tab:mj_flows}
\centering
\adjustbox{max width=0.98\linewidth}{

\begin{tabular}{@{}lcccccc@{}}
\toprule
                    & \multicolumn{3}{c}{100K frames}                                                     & \multicolumn{3}{c}{200K frames}                                            \\
\cmidrule(lr){2-4}\cmidrule(lr){5-7}
Tasks               & \textbf{SSPG}                      & REDQ-FLOW                          & REDQ      & \textbf{SSPG}                      & REDQ-FLOW                 & REDQ      \\ 
\cmidrule(lr){1-1}\cmidrule(lr){2-4}\cmidrule(lr){5-7}
Invertedpendulum-v2 & \textbf{1000±0 (0\%)}              & 868±60 (\red{-13\%})               & 1000±0    & \textbf{1000±0 (0\%)}              & 960±69 (\red{-4\%})       & 1000±0    \\
Hopper-v2           & \textbf{3314±68 (\green{+11\%})}   & 2476±480 (\red{-17\%})             & 2994±510  & \textbf{3487±87 (\green{+14\%})}   & 2940±420 (\red{-4\%})     & 3060±617  \\
Walker2d-v2         & \textbf{4428±230 (\green{+123\%})} & 3477±440 (\green{+75\%})           & 1989±1003 & \textbf{4793±186 (\green{+61\%})}  & 4199±565 (\green{+41\%})  & 2969±861  \\
Halfcheetah-v2      & 8897±496 (\green{+58\%})           & \textbf{8905±1034 (\green{+59\%})} & 5613±436  & \textbf{10309±653 (\green{+55\%})} & 9510±1065 (\green{+43\%}) & 6633±568  \\
Ant-v2              & \textbf{5163±275 (\green{+65\%})}  & 3695±987 (\green{+18\%})           & 3132±1243 & \textbf{5513±238 (\green{+45\%})}  & 4750±795 (\green{+25\%})  & 3792±1064 \\
Humanoid-v2         & \textbf{4992±140 (\green{+256\%})} & 3947±1542 (\green{+181\%})         & 1402±657  & \textbf{5148±51 (\green{+9\%})}    & 4693±238 (\red{-1\%})     & 4721±648  \\ \bottomrule
\end{tabular}

}
\end{table}

In Table~\ref{tab:mj_flows} we provide the per-task comparisons. incorporating flows with REDQ appears to provide limited and inconsistent performance benefits as compared to incorporating our serial Markov chain reasoning framework. This is in line with our intuition justifying SSPG's superiority and with results reported in prior off-policy reinforcement learning work which show flows provide marginal gains over standard Gaussian or deterministic policies (e.g.,~\cite{PG-norm-flows,iterativeAmortizedPolOptim}).

\begin{table}[H]
\tabcolsep=0.05cm
\caption{Per-task results comparing improvements from SSPG and normalizing flows over DrQv2 for the considered DeepMind Control tasks. The results were collected as described in Table~\ref{tab:dmc_perf}. }
\label{tab:dmc_flows}
\begin{center}
\adjustbox{max width=0.98\linewidth}{

\begin{tabular}{@{}lcccccc@{}}
\toprule
                        & \multicolumn{3}{c}{1.5M frames}                                                       & \multicolumn{3}{c}{3.0M frames}                                                    \\
\cmidrule(lr){2-4}\cmidrule(lr){5-7}
Tasks                   & \textbf{SSPG}                             & DrQv2-FLOW                       & DrQv2           & \textbf{SSPG}                            & DrQv2-FLOW                     & DrQv2           \\ 
\cmidrule(lr){1-1}\cmidrule(lr){2-4}\cmidrule(lr){5-7}
Acrobot swingup         & 218±49 (\red{-20\%})             & \textbf{296±16 (\green{+9\%})}   & 272±40          & 371±41 (\red{-12\%})            & 398±37 (\red{-5\%})            & \textbf{422±48} \\
Cartpole swingup sparse & \textbf{797±43 (\green{+67\%})}  & 622±360 (\green{+30\%})          & 478±391         & \textbf{837±15 (\green{+67\%})} & 633±366 (\green{+26\%})        & 503±411         \\
Cheetah run             & 755±47 (\red{-3\%})              & 743±58 (\red{-5\%})              & \textbf{781±32} & \textbf{888±10 (\green{+2\%})}  & 849±51 (\red{-3\%})            & 873±55          \\
Finger turn easy        & 794±127 (\green{+5\%})           & \textbf{834±120 (\green{+10\%})} & 757±156         & \textbf{974±6 (\green{+5\%})}   & 946±50 (\green{+2\%})          & 932±43          \\
Finger turn hard        & \textbf{637±138 (\green{+26\%})} & 555±222 (\green{+10\%})          & 506±229         & \textbf{945±42 (\green{+4\%})}  & 866±60 (\red{-5\%})            & 913±60          \\
Hopper hop              & \textbf{246±28 (\green{+23\%})}  & 172±103 (\red{-14\%})            & 200±102         & \textbf{344±28 (\green{+44\%})} & 217±131 (\red{-9\%})           & 239±123         \\
Quadruped run           & 570±22 (\green{+42\%})           & \textbf{605±64 (\green{+51\%})}  & 402±213         & \textbf{760±64 (\green{+54\%})} & 734±20 (\green{+48\%})         & 494±288         \\
Quadruped walk          & \textbf{855±23 (\green{+45\%})}  & 831±52 (\green{+41\%})           & 591±271         & 888±22 (\red{-2\%})             & \textbf{912±23 (\green{+1\%})} & 905±44          \\
Reach duplo             & \textbf{221±7 (\green{+1\%})}    & 218±7 (\red{-0\%})               & 219±7           & 218±9 (\red{-4\%})              & 227±1 (\red{-0\%})             & \textbf{228±1}  \\
Reacher easy            & \textbf{978±4 (\green{+1\%})}    & 976±3 (\green{+0\%})             & 973±3           & \textbf{982±3 (\green{+3\%})}   & 973±16 (\green{+2\%})          & 954±22          \\
Reacher hard            & \textbf{913±77 (\green{+14\%})}  & 895±20 (\green{+12\%})           & 802±113         & \textbf{974±6 (\green{+3\%})}   & 956±18 (\green{+1\%})          & 944±25          \\
Walker run              & \textbf{634±16 (\green{+12\%})}  & 520±287 (\red{-9\%})             & 568±273         & \textbf{738±7 (\green{+20\%})}  & 575±319 (\red{-7\%})           & 616±297         \\ 
\cmidrule(lr){1-1}\cmidrule(lr){2-4}\cmidrule(lr){5-7}
Average score           & \textbf{634.79 (\green{+16\%})}  & 605.41 (\green{+11\%})           & 545.72          & \textbf{743.32 (\green{+11\%})} & 690.45 (\green{+3\%})          & 668.60          \\
\cmidrule(lr){1-1}\cmidrule(lr){2-4}\cmidrule(lr){5-7}
Median score            & \textbf{695.85 (\green{+29\%})}  & 613.28 (\green{+14\%})           & 537.37          & \textbf{862.73 (\green{+16\%})} & 791.32 (\green{+6\%})          & 744.66          \\ \bottomrule
\end{tabular}

}
\end{center}
\end{table}

In Table~\ref{tab:mj_flows} we provide the per-task comparisons. Similarly to our results in the OpenAI Gym Mujoco tasks, incorporating flows with DrQv2 also appears to provide limited and inconsistent benefits as compared to the benefits of SSPG. We note that in both benchmarks, flows appear to particularly help earlier in training (100K steps and 1.5M steps, respectively), likely due to some initial exploration benefits. However, its gains over DrQv2 after 3M steps remain marginal in 10 out of the 12 examined DeepMind Control tasks.

\begin{figure}[H]
    \centering
    \includegraphics[width=0.8\linewidth]{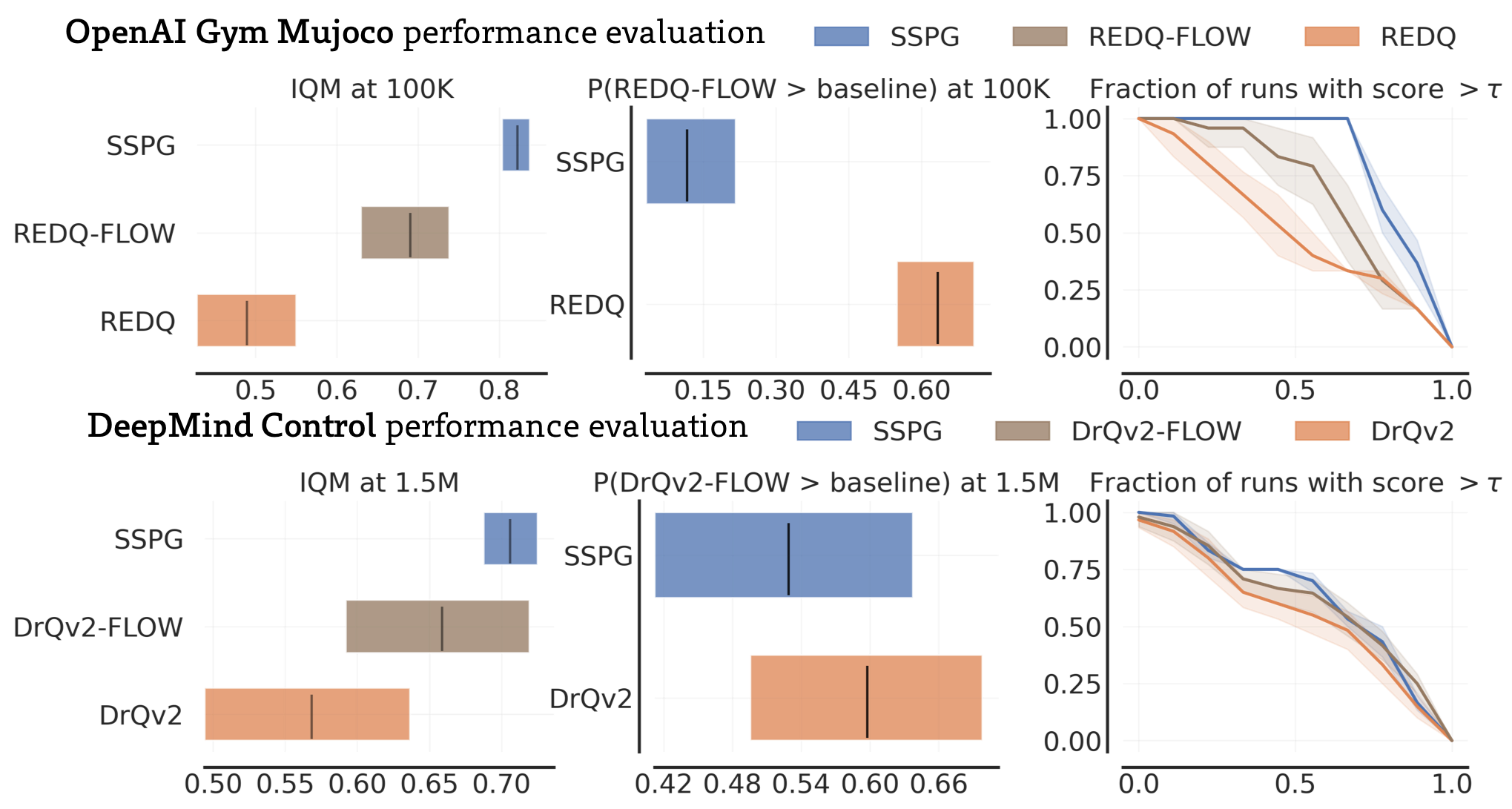}
    \caption{Performance evaluation of our normalizing flow agents with \textit{Rliable} \citep{precipice-rliable}. We integrate flows with REDQ for the OpenAI Gym Mujoco tasks (\textbf{Top}) and  with DrQv2 for the DeepMind Control tasks from pixels   (\textbf{Bottom}).}
    \label{fig:app:ext_rliable_flows}
\end{figure}%

In Figure~\ref{fig:app:ext_rliable_flows}, we provide aggregate metrics for the normalizing flows agents, using the same statistical tools described in Section~\ref{sec:4exps}. We considered performance after 100K and 1.5M steps for the OpenAI Gym Mujoco and DeepMind Control benchmarks, since its where normalizing flows appear to yield the most benefits. However, even in this experience regime, incorporating flows does not provide statistically significant gains and does not exhibit stochastic dominance \citep{stoch-dom} as SSPG.

\subsection{Computational requirements}

\begin{table}[H]
\caption{Average training times for the tested algorithms and ablations in the OpenAI Gym and DeepMind Control tasks.}
\label{tab:train_time}
\begin{center}
\begin{tabular}{lc} 
\toprule
OpenAI Gym Mujoco                & Training time (seconds) for 1000 env. steps   \\ 
\midrule
SSPG (UTD=10)                    & 50.9                                      \\ 
SSPG (UTD=20)                    & 84.2                                    \\
SSPG, 1 reasoning step (UTD=20)  & 69.1                                    \\ 
SAC-20                           & 51.6                                      \\
IAPO (Original implementation)   & 193.1                                     \\
REDQ (Original implementation)   & 183.8                                   \\ 
\midrule
DeepMind Control (pixels)        & Training time (seconds) for 10000 env. steps  \\ 
\midrule
SSPG (UTD=0.5)                   & 160.2                                   \\ 
SSPG, 1 reasoning step (UTD=0.5) & 116.7                                     \\ 
DrQv2                            & 111.2                                   \\
\bottomrule
\end{tabular}
\end{center}
\end{table}

We record the average training time from executing the different considered algorithms for each \textit{training epoch}, consisting of 1000 environment steps for OpenAI Gym and 10000 environment steps for DeepMind Control. The most relevant specifications of our hardware setup are an \textit{NVIDIA RTX 3090} GPU and an \textit{AMD Ryzen Threadripper 3970x} CPU. We based our implementations on the shared code from \citep{learningpessimism} and \citep{drqv2} for OpenAI Gym and DeepMind Control tasks, respectively. 

\textbf{Training times.} We show the average training times in Table~\ref{tab:train_time}. For the OpenAI Gym Mujoco tasks, the increased UTD ratio severely affects computation requirements. Thus, by halving the UTD ratio of SSPG, we are able to bring its epoch training time lower than SAC-20 and all other considered baselines. However, we did not explore lowering the number of updates for our experiments in DeepMind Control, as DrQv2 already employs a low UTD ratio of 0.5. Thus, DrQv2 is about 30\% computationally faster than SSPG. We believe the extra computation for carrying out the reasoning process is one of the main limitations of the \textit{serial Markov chain reasoning} framework. However, for real-world problems, physical constraints related to environment feedback are often the main bottlenecks for the overall time efficiency. In such regard, we argue that the considerable sample-efficiency improvements from our novel framework should be much more relevant than increased computational cost.

\subsection{Analysis of the cost of acting during deployment}

\begin{table}[H]
\tabcolsep=0.12cm
\caption{Performance of SSPG after `clipping' the maximum number of reasoning steps during evaluation only. We write the number of clipping steps after the algorithm's name.} 
\label{tab:analysis_action_cost}
\centering
\adjustbox{max width=0.98\linewidth}{\begin{tabular}{lcccccc} 
\toprule
Tasks      &SSPG       & SSPG-1                    &SSPG-4                         &SSPG-8                             &REDQ \\ \midrule
Ant-v2              &5513±238   &5356±261                   &5501±114                       &5527±219                           &3792±1064\\
Humanoid-v2         &5148±51    &4971±98                    &5120±56                        &5139±76                            &4721±648 \\ \midrule
Tasks      &SSPG       & SSPG-1                    &SSPG-4                         &SSPG-8                             &DrQv2 \\ \midrule
cheetah\_run        &888±10     &880±13                     &889±10                         &891±5                              &873±55  \\
quadruped\_run      &760±64     &715±85                     &734±81                         &752±66                             &494±288 \\ 
\bottomrule
\end{tabular}}
\end{table}

We report the performance of SSPG when `clipping' the maximum number of reasoning steps allowed for each action-selection to fixed values, during evaluation only. We evaluate final checkpoints of agents learned without any clipping by our unmodified SSPG. We consider a subset of four total environments from both OpenAI Gym Mujoco (i.e., Humanoid-v2, and Ant-v2) and DeepMind Control (i.e., cheetah\_run, and quadruped\_run) for which SSPG displays different average reasoning requirements. As shown in Table~\ref{tab:analysis_action_cost}, clipping to as low as four reasoning steps marginally affects the performance of SSPG, which still always surpasses the scores achieved by standard reinforcement learning baselines. SSPG is less affected by this form of deployment-only clipping than by fixing the number of reasoning steps for both training and evaluation phases. We motivate this finding by noting that better capturing the canonical distribution of returns from the critic by performing additional reasoning steps has also significant benefits during the training phase of off-policy algorithms. In particular, this allows the agent to achieve better exploration and more easily correct the critic in the areas of the action space where its predictions are erroneously optimistic. When training with the unmodified SSPG algorithm, this benefit is fully retained, justifying the performance superiority than our previous fixed-steps ablations.

\begin{table}[H]
\caption{Average deployment times for the tested algorithms and ablations in the OpenAI Gym and DeepMind Control tasks.}
\label{tab:deploy_time}
\begin{center}
\begin{tabular}{lc} 
\toprule
OpenAI Gym Mujoco                & Deployment time (seconds) for 1000 env. steps   \\ 
\midrule
Random (only simulation)         & 0.583                    \\
SSPG                             & 3.523                     \\
SSPG-1 (clipping)                           & 2.462                     \\
SSPG-4 (clipping)                           & 2.917                     \\ 
SSPG-8 (clipping)                           & 3.257                     \\
SAC-20                           & 2.451                     \\ 
IAPO (Original implementation)   & 5.657                     \\
REDQ (Original implementation)   & 2.671                     \\ 
\midrule
DeepMind Control (pixels)        & Deployment time (seconds) for 10000 env. steps  \\ 
\midrule
Random (only simulation)         & 0.997                     \\
SSPG                             & 1.701                     \\
SSPG-1 (clipping)                           & 1.637                     \\ 
SSPG-4 (clipping)                           & 1.693                     \\ 
SSPG-8 (clipping)                           & 1.699                     \\ 
DrQv2                            & 1.604                     \\ 
\bottomrule
\end{tabular}
\end{center}
\end{table}

We report the average rollout time during deployment of each of our implementations. As shown in Table~\ref{tab:deploy_time}, using SSPG does increase the average rollout time over standard reinforcement learning baselines. However, the additional time required for action-selection scales sub-linearly with the number of reasoning steps, and appears to be dominated by other fixed costs, such as simulating the environment and converting observations to tensor objects. This is in contrast with the other more expensive iterative baseline, IAPO~\citep{iterativeAmortizedPolOptim}, which performs gradient-based optimization at each acting step. Differences with standard reinforcement learning are even more marginal in the visual DeepMind Control environments, where the most expensive part of the computation is from encoding the RGB input observation with a convolutional encoder (which needs to occur only once before performing the reasoning steps with the policy `head'). Moreover, we note that in many real-world applications, the additional acting cost would be greatly inferior to the actuation time costs when using distributed hardware. However, clipping the number of reasoning steps still remains a valuable option, as examined in the experiment above.  

\newpage

\section{Ablations and parameter studies}

\label{app:abl}

We perform additional experiments to evaluate SSPG's design choices and test alternative optimization setups for \textit{serial Markov chain reasoning}. We focus on the Humanoid-v2 and quadruped\_run tasks, which we found to be generally representative of overall agent behavior in the OpenAI Gym Mujoco and DeepMind Control domains, respectively. For each experiment, we report the performance curves for average returns (higher is better) and number of reasoning steps (lower is better).

\subsection{Loss backpropagation}

\begin{figure}[H]
    \centering
    \includegraphics[width=0.85\linewidth]{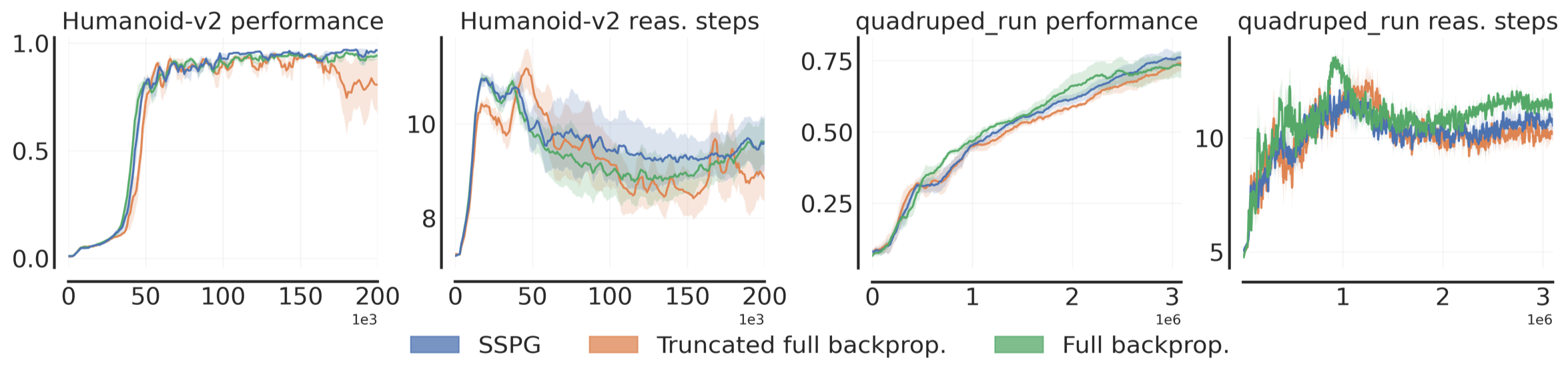}
    \caption{Performance and reasoning steps with alternative backpropagation strategies for training SSPG's BT-policy.}
    \label{fig:app:abl_bp}
\end{figure}%

Following the policy gradient estimation derived in Section~\ref{subsec:learning_bt_pol}, we optimize the BT-policy by recording its gradients with respect to the following operations: 1) We perform one (reparameterized) reasoning step to output an action-belief, $a_0\sim \pi^b_\theta(s, a)$. 2) We simulate a \textit{local approximation} of the (reparameterized) RMC for $N$ steps, yielding $a_{1:N}$. 3) We compute and backpropagate the outputs of all $Q^s_\phi(s, a_n)$, for $n=0,...,N$. A practical alternative would be to carry out step (2) by performing $N$ reasoning steps with the (reparameterized) BT-policy itself, rather than with a local approximation of the RMC. While this change would not affect the objective value from each $Q^s_\phi(s, a_n)$, it would make the computation record additional gradient dependencies to the BT-policy's parameters \textit{from each reasoning step}, rather than just the first. As we already have to perform forward and backward passes for the $N$ actions computed in step (2), this extension would not come at a significant computational overhead and would be reminiscent of differentiation through a traditional recurrent model. In practice, we expect this change would enable to reduce optimization variance, as we are optimizing the \textit{direct influence} that the BT-policy has on the $RMC$ starting from $N$ different initial action-beliefs for each sample. Yet, this alternative optimization procedure is not supported by our theoretical intuition and would likely introduce further bias into the policy gradient estimation.

We show experimental results in Figure~\ref{fig:app:abl_bp}. We evaluate both a truncated version (truncated full backprop.) and an un-truncated version (full backprop.) that perform \textit{full backpropagation} through the RMC when optimizing SSPG. The truncated version still optimizes the BT-policy with respect to the full $N$ reasoning steps, but prevents the gradients of future steps backpropagating to earlier action-beliefs (somewhat reminiscent of truncation in RNN optimization). Surprisingly, we observe no major differences performance-wise. However, we do observe somewhat higher variance between different experiments when performing full backpropagation and slightly lower final returns, likely symptomatic of the additional bias. On the other hand, the un-truncated version does experience marginally faster initial learning in quadruped\_run, indicating there could be some practical benefits from backpropagating multiple reasoning steps to the BT-policy's parameters. We leave exploration of such potential extensions for future work.

\subsection{Initial action beliefs}

\begin{figure}[H]
    \centering
    \includegraphics[width=0.85\linewidth]{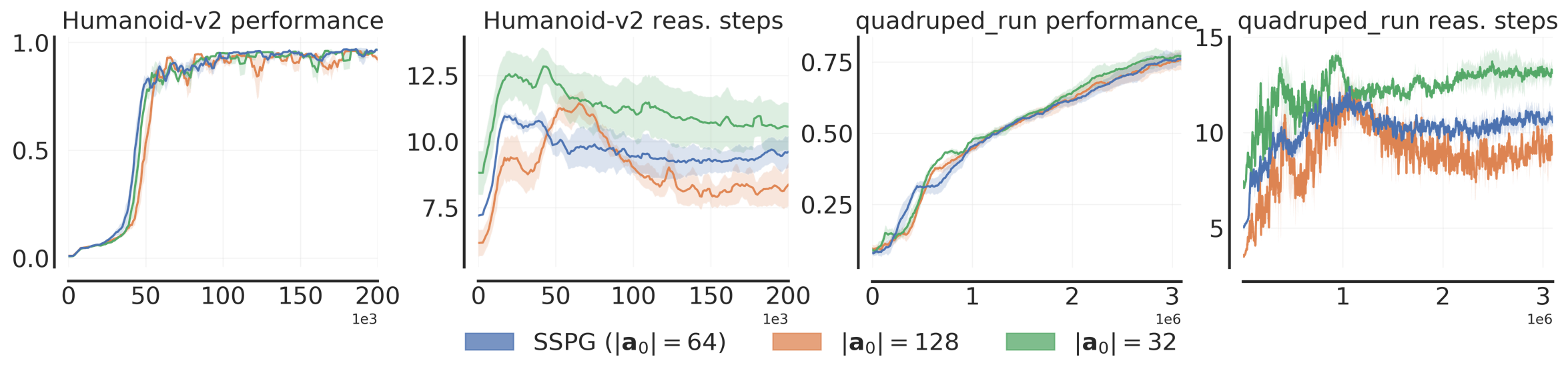}
    \caption{Performance and reasoning steps with double and half SSPG's number of initial action-beliefs.}
    \label{fig:app:abl_ini_actbel}
\end{figure}%

As described in Section~\ref{subsec:action_sel_temp_cons}, we perform the reasoning process  from a set of initial action-beliefs $\mathbf{a}_0$. The size of $\mathbf{a}_0$ (denoted $|\mathbf{a}_0|$) influences the number of initial starting modes and the accuracy of the empirical PSRF computation for convergence detection. By default SSPG employs $|\mathbf{a}_0|=64$. As shown in Figure~\ref{fig:app:abl_ini_actbel}, doubling or halving $|\mathbf{a}_0|$ affects convergence speed, with a greater number of initial action-beliefs leading to a slightly faster reasoning process. We attribute this phenomenon to the conservativeness of the proposed PSRF convergence rule \citep{gelmanrubin-revisited}, which becomes harder to satisfy when evaluated from fewer parallel chains. However, these results do not take into account the cost of performing each reasoning step which is itself affected by $|\mathbf{a}_0|$. SSPG's overall efficiency should be quite robust to most choices for this hyper-parameter with hardware optimized for parallel computation.

\subsection{Short term action memory buffer size}

\begin{figure}[H]
    \centering
    \includegraphics[width=0.85\linewidth]{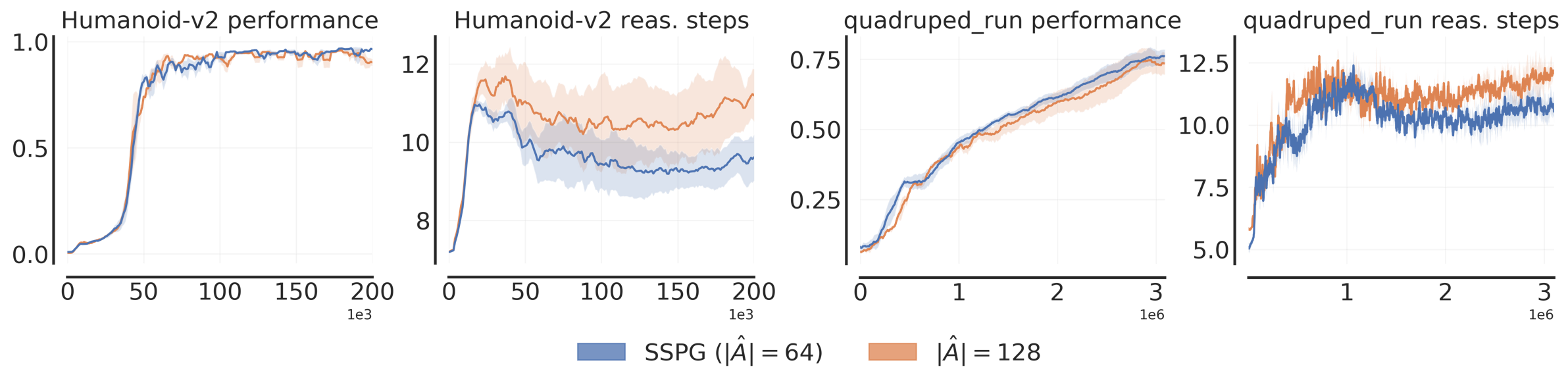}
    \caption{Performance and reasoning steps using a \textit{short-term action memory} buffer with increased size.}
    \label{fig:app:abl_actbuff}
\end{figure}%

As described in Section~\ref{subsec:action_sel_temp_cons}, we sample initial action-beliefs for reasoning from the \textit{short-term action memory} buffer, $\hat{A}$. The size of $\hat{A}$ (denoted $|\hat{A}|$) determines how far into the past to store action-beliefs that can facilitate future reasoning. In practice, we found that the simple heuristic of setting $|\hat{A}|=|\mathbf{a}_0|$ works well. In Figure~\ref{fig:app:abl_actbuff}, we examine the effects of doubling the buffer size from 64 to 128. Again, we observe this change leads to an increase in the expected reasoning steps before reaching convergence. This result indicates that the most recent past actions are generally most relevant, which we believe is a consequence of the fine temporal discretization of the considered environments.

\newpage

\section{Additional supporting analysis results}

\label{app:add_ana} %

\subsection{Positional bandits quantized transition probabilities}

\begin{figure}[H]
    \centering
    \includegraphics[width=0.8\linewidth]{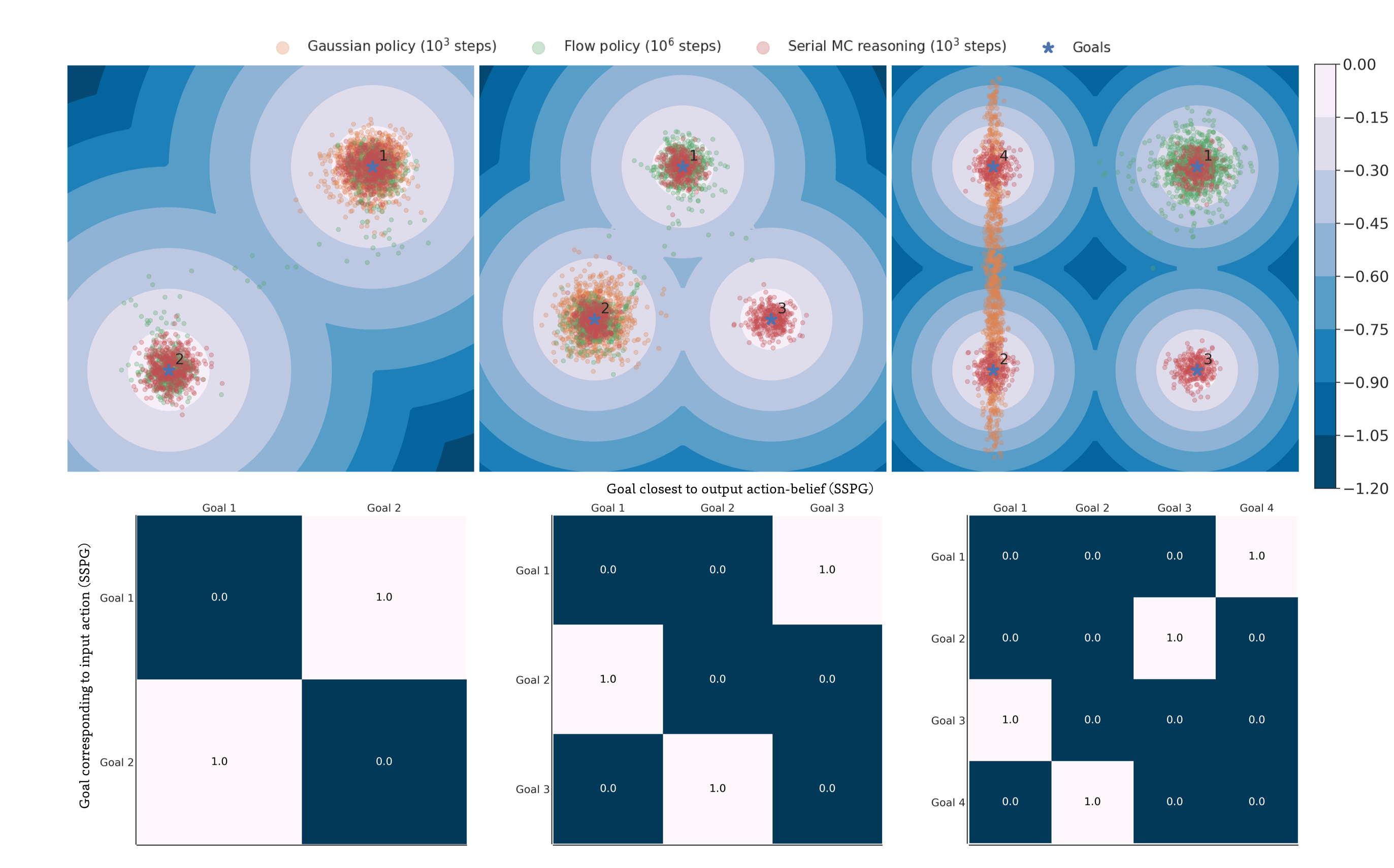}
    \caption{Quantized transition matrices visualizations for the learned BT-policy trained in the positional bandits with 2, 3, and 4 goals (see Figure~\ref{fig:ana_comp} A).}
    \label{fig:app:ext_quant}
\end{figure}%

We compute the quantized transition probabilities of the RMCs in the considered positional bandits from our \textit{Policy expressiveness} analysis (Section~\ref{subsec:4prop_exps}). For each positional bandit we randomly sample a set of 1000 action-beliefs within a radius around each goal and compute the likelihood of transitioning between any two such action-beliefs. We sum and normalize these probabilities both with respect to the input and output action beliefs within each goal. Hence, we obtain a transition matrix for the discretized RMC, showing the probability of updating the current action-belief based on the resulting closest goal. We find that the BT-policy intuitively learns to consider action-beliefs to reach all different goals in turn. For instance, in the positional bandit with three goals, performing a reasoning step with the learned BT-policy from an action-belief that would land nearby goal 1 would almost certainly lead to an action-belief that would land near goal 3. Similarly, an action-belief landing near goal 2 would follow from an action-belief that would land near goal 3, and would almost certainly lead to an action-belief that would land near goal 1 (with less than $10^{-6}$ probability of deviating). This behavior is summarized from visualizing the relative transition matrices following the described quantization in Figure~\ref{fig:app:ext_quant}. 

\subsection{Reasoning with fixed action beliefs}

\begin{figure}[H]
    \centering
    \includegraphics[width=0.85\linewidth]{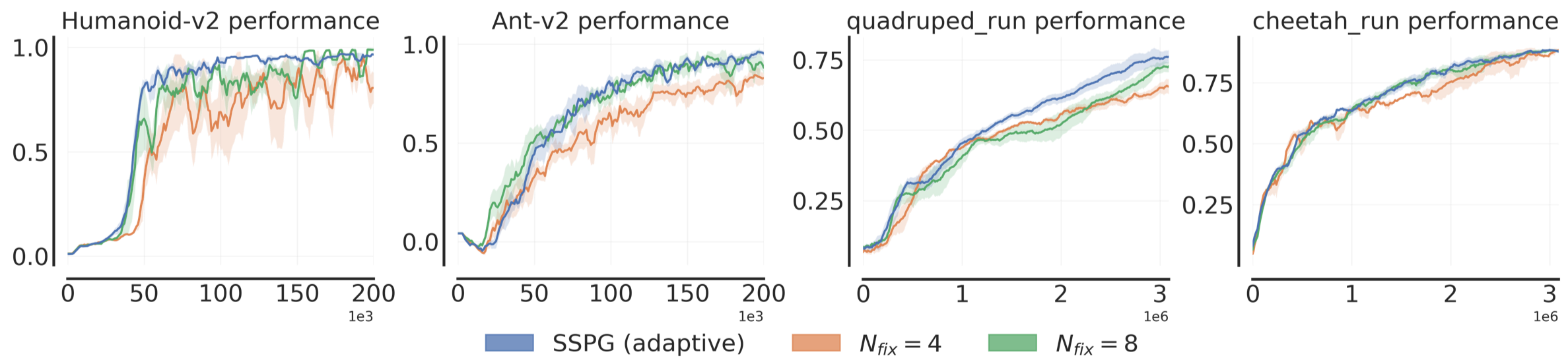}
    \caption{Performance from performing reasoning with alternative \textit{fixed} number of steps in different tasks. The average number of reasoning steps until convergence with SSPG for Humanoid-v2 quadruped\_run is around 10, while for Ant-v2 and cheetah\_run is around 5.}
    \label{fig:app:ext_fix}
\end{figure}%

We provide further results in support of our \textit{Policy adaptivity} analysis by comparing the performance of our adaptive SSPG with performing alternative fixed numbers of reasoning steps in 4 different tasks with diverse average reasoning costs (see Figure~\ref{fig:ana_comp} B). As shown in Figure~\ref{fig:app:ext_fix} and also Figure~\ref{fig:ana_comp} C, increasing the number of reasoning steps monotonically improves performance. Yet, we did not observe any case where our adaptive strategy underperforms as compared to any fixed number of steps. We believe this is further empirical evidence of how a framework that can adaptively dedicates different amounts of computation to different action-selection problem can provide great efficiency and scaling benefits.

\subsection{Effects of short-term action memory buffer}

\begin{figure}[H]
    \centering
    \includegraphics[width=0.85\linewidth]{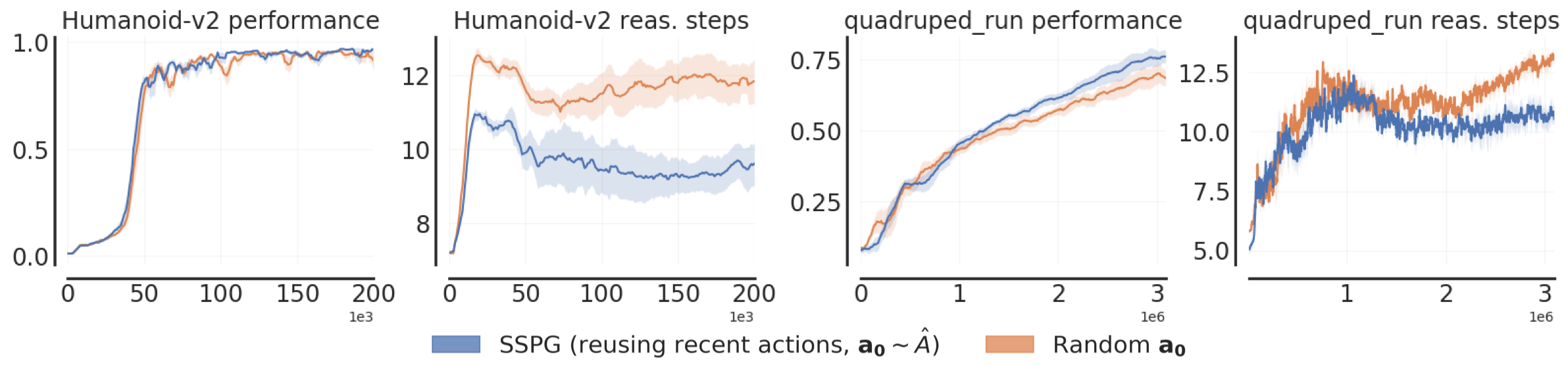}
    \caption{Performance and reasoning steps with and without reusing recent actions as initial action-beliefs.}
    \label{fig:app:ext_actmem}
\end{figure}%

As stated in the \textit{Solution reuse} analysis, and shown in Figure~\ref{fig:app:ext_actmem}, re-initializing randomly $\mathbf{a}_0$ hurts efficiency but not performance. We believe this is an indication that the BT-policy effectively learns to \textit{bootstrap} information in each previous action-belief while reasoning. Intuitively, starting from an initial action-belief that contains no information about optimal behavior makes the relative action-selection problem harder. Yet, SSPG still successfully enables to recover very similar performance via adaptively performing additional reasoning steps to compensate.

\section{Societal impact}

\label{app:soc_imp}

We proposed a new framework for modeling autonomous decision-making in reinforcement learning. Thus, the societal impact of our work is tightly related to the impact of the reinforcement learning field. While autonomous systems can offer many benefits, as they become more commonplace, they might introduce ethical issues such as privacy, surveillance, and bias. If unregulated, automation might also accentuate economic inequalities and have non-trivial environment impact. Policy making strategies and regulations are increasingly needed to mitigate these risks, which we believe should not compromise the field's advancements given its countless potential positive implications.

\end{document}